\pdfoutput=1
\documentclass[11pt]{article}

\usepackage[final]{acl}

\usepackage{times}
\usepackage{latexsym}
\usepackage{booktabs}
\usepackage{amssymb}
\usepackage[T1]{fontenc}
\usepackage[utf8]{inputenc}

\usepackage{microtype}

\usepackage{inconsolata}

\usepackage{graphicx}
\usepackage{amsmath}
\usepackage[hang,flushmargin]{footmisc}
\usepackage{listings}
\usepackage{arydshln}
\usepackage{xcolor}

\usepackage{wrapfig}
\usepackage{cleveref}
\usepackage[linesnumbered, ruled, vlined]{algorithm2e}

\lstdefinestyle{pythonstyle}{
    language=Python,
    basicstyle=\ttfamily\tiny,
    keywordstyle=\color{blue},
    commentstyle=\color{green!50!black},
    stringstyle=\color{red},
    numberstyle=\tiny\color{gray},
    stepnumber=1,
    breaklines=true,
    breakatwhitespace=true,
    tabsize=4,
    showstringspaces=false,
    frame=single
}

\newcommand{\ourmethod}{\textsc{LaSP}\xspace}

\title{Language-to-Space Programming for Training-Free 3D Visual Grounding}

\usepackage{xspace}
\newcommand{\eg}{e.g.\@\xspace}

\author{Boyu Mi$^{1, 2}$ \quad Hanqing Wang$^2$ \quad Tai Wang$^2$ \quad Yilun Chen$^2$ \quad Jiangmiao Pang$^2$\\   \vspace{4pt}\\
$^1$Shanghai Jiao Tong University \quad $^2$Shanghai Artificial Intelligence Laboratory \\
\texttt{miboyu@pjlab.org.cn}
}

\begin{document}
\maketitle
\begin{center}
    \begin{figure*}
    \includegraphics[width=\linewidth]{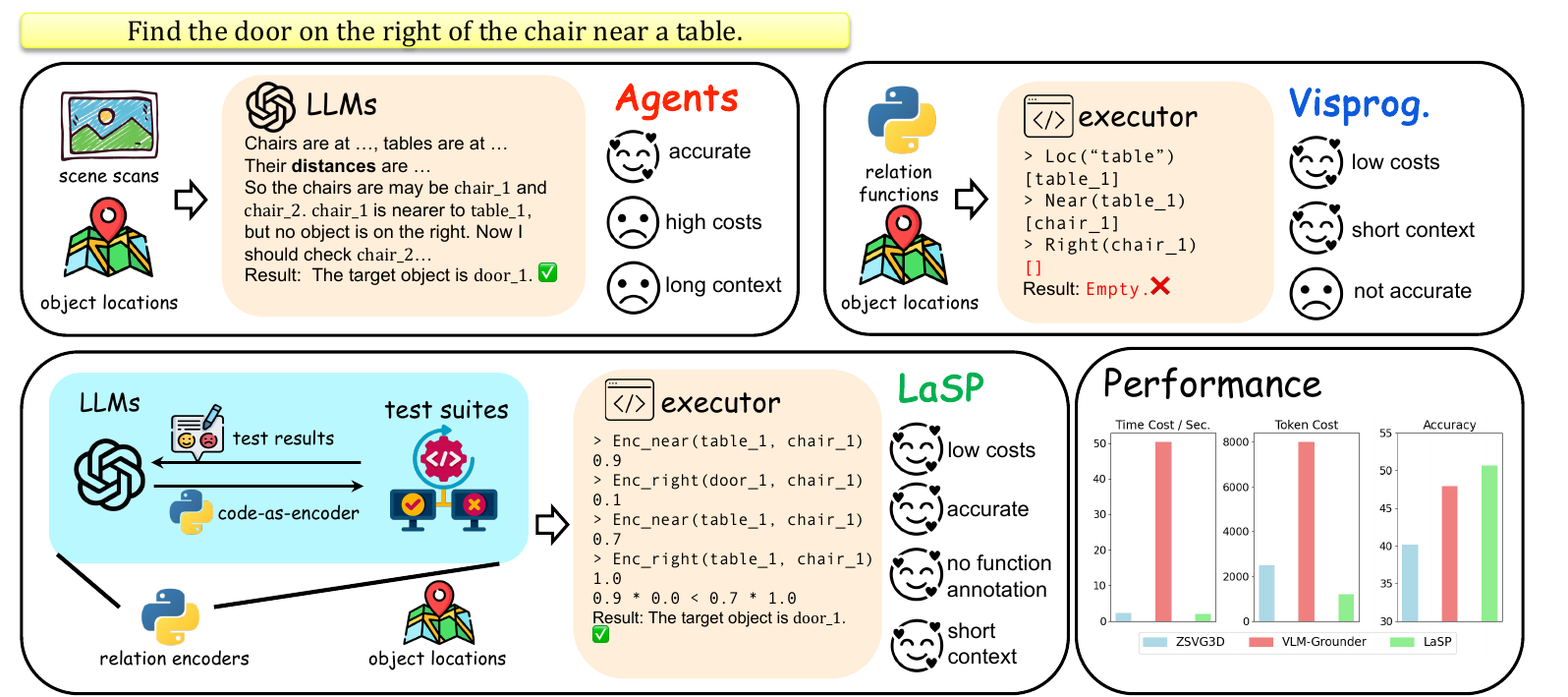}
    \captionof{figure}{
    Accuracy and cost comparison of \textcolor{green}{\ourmethod} (ours) with two types of existing training-free 3DVG methods.
    \textcolor{red}{Agent}-based methods input scene information into LLMs/VLMs to analyze spatial relations, leading to high accuracy but also high computational costs.
    Visual programming (\textcolor{blue}{Visprog.}) only inputs the referring utterance into LLMs to generate a program and finds the target by program execution. It reduces the costs signicicantly but sacrifices the accuracy.
    \textcolor{green}{\ourmethod} introduces code-based relation encoders along with its automatic generation pipeline.
    Spatial relations are analyzed by code execution instead of LLMs/VLMs reasoning. This approach allows \ourmethod to achieve accuracy comparable to agent-based methods, while significantly reducing the costs.
    } 
    \label{fig:banner}
    \end{figure*}
\end{center}
\begin{abstract}

3D visual grounding (3DVG) is challenging due to the need to understand 3D spatial relations.
While supervised approaches have achieved superior performance, they are constrained by the scarcity and high annotation costs of 3D vision-language datasets.
Training-free approaches based on LLMs/VLMs eliminate the need for large-scale training data, but they either incur prohibitive grounding time and token costs or have unsatisfactory accuracy.
To address the challenges, we introduce a novel method for training-free\footnote{
In this paper, we define ``training-free'' as an approach that aims to eliminate the dependency on large-scale, expensively annotated vision-language datasets. Although our method leverages a small set of examples to optimize its components, its core contribution aligns with that of other zero-shot methods. We therefore classify it as such.} 3D visual grounding, namely
\underline{La}nguage-to-\underline{S}pace \underline{P}rogramming (\ourmethod).
\ourmethod introduces LLM-generated codes to analyze 3D spatial relations among objects, along with a pipeline that evaluates and optimizes the codes automatically.
Experimental results demonstrate that \ourmethod achieves 52.9\% accuracy on the Nr3D benchmark, ranking among the best training-free methods.
Moreover, it substantially reduces the grounding time and token costs, offering a balanced trade-off between performance and efficiency.
Code is available at \url{https://github.com/InternRobotics/LaSP}.
\end{abstract}

\section{Introduction}
\label{sec:intro}
The 3D visual grounding (3DVG) task focuses on locating an object in a 3D scene based on a referring utterance~\cite{liu2024survey}.
Numerous supervised methods have been proposed for 3DVG~\cite{hsu2023ns3d, jain2022bottom, huang2022multi, chen2022language, huang2024chat, 3dvista, bakr2023cot3dref, wu2023eda}. 
These methods learn representations of referring utterances, object attributes, and spatial relations from large-scale 3D vision-language training datasets with high-quality annotations and achieve state-of-the-art performance on 3DVG.
However, the scarcity of 3D vision-language datasets~\cite{chen2020scanrefer, achlioptas2020referit3d}, coupled with the high cost of their annotations, limits these methods' applicability.
% Furthermore, some supervised methods are trained on these closed-vocabulary datasets, restricting them in open-vocabulary scenarios.

Recently, large language models (LLMs) and vision-language models (VLMs) have shown remarkable capabilities in reasoning, code generation, and visual perception. 
Building on these advancements, open-vocabulary and zero-shot methods~\cite{yang2024llm, xuvlm, fang2024transcrib3d, yuan2021instancerefer, li2024seeground} are proposed.
Agent-based methods~\cite{yang2024llm, xuvlm, fang2024transcrib3d} always let LLMs perform numerical reasoning on object locations and in text modality to find the target object~\cite{yang2024llm, fang2024transcrib3d}, or let VLMs locate targets from scene scan images in visual modality~\cite{xuvlm}.
These agents achieve superior accuracy compared to other training-free methods, but for one referring utterance, they need to input the whole scene information into LLMs/VLMs. 
Before finding the target object, LLMs/VLMs always generate lengthy responses, containing planning, reasoning, or self-debugging processes.
This results in high costs in terms of grounding time and token usage (see Figure~\ref{fig:banner},~\textcolor{red}{\texttt{Agents}} block).
In contrast, the visual programming method~\cite{yuan2024visual} only inputs the referring utterance into LLMs to generate a short program which calls annotated selection functions. Then the program execution, which is much faster than LLM reasoning, outputs the target object. As a result, its time and token costs are much lower.
However, it has trouble considering multiple spatial relations in the referring utterance simultaneously~\cite{csvg}, resulting in relatively low accuracy.(see Figure~\ref{fig:banner},~\textcolor{blue}{\texttt{Visprog.}} block.)

To address the dual challenges of accuracy and costs, we propose \textbf{\ourmethod}, a novel training-free 3DVG method that balances the accuracy and grounding costs.
(see Figure~\ref{fig:banner},~\textcolor{green}{~\texttt{LaSP}} block.)
Specifically, 
\ourmethod uses Python codes that are generated and optimized by LLMs as spatial relation encoders.
Given the bounding boxes of scene objects, the spatial relation encoders generate relation features which quantify the spatial relations of objects.
Moreover, we introduce test suites which can evaluate the codes.
The test suites not only enable us to select better relation encoders from multiple LLM responses but also allow LLMs to leverage failed test cases to optimize the codes.
The relation encoders can be seamlessly integrated with a symbolic reasoning framework
similar to~\cite{hsu2023ns3d}.
In our framework, a referring utterance is converted to a symbolic expression.
Then an executor aggregates the symbolic expression, relation features, and object categories to give the matching scores between objects and the referring utterance.
\ourmethod also prompts VLMs to further distinguish objects based on visual information.
Compared to agent-based methods, \ourmethod only inputs the referring utterance into LLMs and one image into VLMs, resulting in much lower costs.
Compared to the visual programming method, \ourmethod has obviously higher accuracy.

We evaluate \ourmethod on the widely used Nr3D~\cite{achlioptas2020referit3d} datasets.
Experiment results show that {\ourmethod} achieves 52.9\% accuracy on Nr3D, and offers advantages in grounding time and token cost compared to previous training-free 3D visual grounding methods.
Additionally, we conduct experiments to demostrate the advantages the LLM-desgined codes over human experts and the generalization to other 3D datasets.
% {\ourmethod} can outperform VLMGrounder~\cite{xuvlm} on Nr3D in accuracy while requiring less than $1/5$ of the grounding time and less than $1/2$ of the token usage.
% In addition, {\ourmethod} significantly outperforms ZSVG3D~\cite{yuan2024visual} in accuracy with comparable grounding time and token costs (see~\cref{fig:banner},~\texttt{Grounding Performance} block). 
% The evaluation results show that 
% among various training-free 3DVG methods, {\ourmethod} strikes an excellent balance between accuracy and efficiency.

\begin{figure*}[h]
    \centering
    \includegraphics[width=\linewidth]{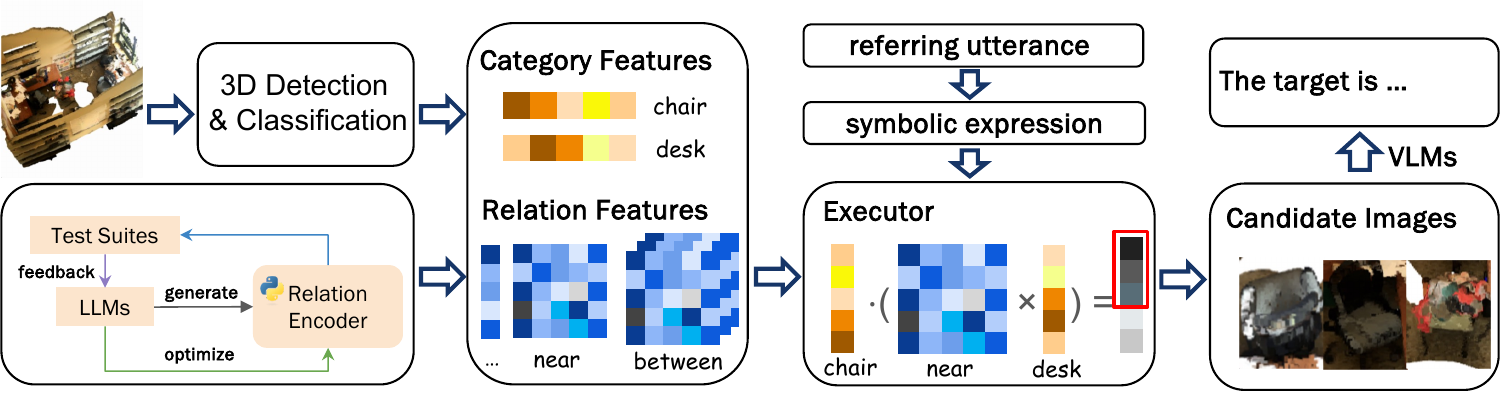}
    \caption{\textbf{Overview of \ourmethod}.
    Off-the-shelf spatial relation encoders are generated and optimized by LLMs before grounding.
    At the grounding time, the encoders compute relation features based on object bounding boxes.
    An executor uses the relation features, along with category features and the symbolic expression to get some candidate objects. Then \ourmethod uses VLMs to select the target from their images.
    }
    \label{fig:framework}
\end{figure*}

\section{Related Work}
\label{sec:related}
% \paragraph{Supervided 3D Visual Grounding}
% 3D visual grounding (3DVG) aims to localize objects in 3D scenes based on natural language descriptions of appearance and spatial relations. 
% Two dominant benchmarks, ScanRefer~\cite{chen2020scanrefer} and ReferIt3D~\cite{achlioptas2020referit3d}, leverage ScanNet~\cite{dai2017scannet} scenes to provide diverse object-utterance pairs. 
% Supervised methods typically train end-to-end models on annotated 3D vision-language datasets. 
% BUTD-DETR\cite{jain2022bottom} integrates bottom-up object detection with transformer-based grounding, 
% ViL3DRel\cite{chen2022language} designs fine-grained neural networks to encode spatial relations.
% Though achieving promising accuracy, these methods suffer from expensive data annotation dependency~\cite{3dvista}.
% Neuro-symbolic methods~\cite{hsu2023ns3d, feng2024naturally, li2024r2g} attempt to mitigate data reliance by combining symbolic parsing with neural components. 
% They parse referring utterances into symbolic expressions using LLMs and train neural networks as spatial relation encoders.
% Unlike these approaches, \ourmethod avoids the need for specific 3D training.
\paragraph{Training-free 3D Visual Grounding}
Training-free methods exploit pre-trained LLMs / VLMs for open-vocabulary 3DVG. 
ZSVG3D~\cite{yuan2024visual} uses LLMs to generate programs that call predefined functions to find the target object. CSVG~\cite{csvg} proposes to replace the programming of ZSVG3D~\cite{yuan2024visual} by constraint satisfaction solving for handling multiple constraints.
LLM-Grounder~\cite{yang2024llm}, Transcrib3D~\cite{fang2024transcrib3d} deploy LLM/VLM-based agents that analyze object appearances and locations and find the target.
VLM-Grounder~\cite{xuvlm} and SeeGround~\cite{li2024seeground} mainly rely on VLMs to find the target from scene images by visual prompting.
\citet{xuvlm} uses VLMs and images from the scene to figure out the target object. \citet{li2024seeground} first parses the landmark and perspective of the referring utterance and then uses VLMs to find the target object from rendered images.
Compared to these methods, \ourmethod offers a superior results on both accuracy and efficiency.

\paragraph{LLM-based Code Generation}
LLMs demonstrate growing proficiency in generating executable code~\cite{roziere2023code} for precise mathematical reasoning~\cite{li2023chain}, robotics control~\cite{liang2023code}, tool use~\cite{gupta2023visual, yuan2024visual} or data cleaning~\cite{zhou2024programming}.
Recent work further explores code refinement via environmental feedback, such as RL training trajectories~\cite{ma2023eureka} or real-world execution errors~\cite{le2022coderl, chen2023teaching}.
In the 3DVG area, \cite{yuan2024visual, fang2024transcrib3d} also uses code to process spatial relations, but \ourmethod advances this paradigm by introducing the spatial relation encoders and test suites to automatically optimize codes.

\section{Method}
\label{sec:method}
\subsection{Problem Statement}

3D visual grounding tasks involve a scene, denoted as $\mathcal{S}$, represented by an RGB-colored point cloud containing $C$ points. Associated with this is an utterance $\mathcal{U}$ that describes an object within the scene $\mathcal{S}$. The objective is to identify the location of the target object $\mathcal{T}$ in the form of a 3D bounding box. In the ReferIt3D dataset~\cite{achlioptas2020referit3d}, bounding boxes for all objects are provided, making the visual grounding process a task of matching these bounding boxes to the scene $\mathcal{S}$.
% In contrast, the ScanRefer dataset~\cite{chen2020scanrefer} provides only the point cloud of the scene, requiring additional detection or segmentation modules to accomplish the grounding task.

\subsection{Grounding Pipeline}
\label{sec:pipeline}

The framework of \ourmethod is shown in Figure~\ref{fig:framework}.
Prior to grounding, relation encoders are generated by LLMs, and objects in 3D scenes are detected and classified. 
A semantic parser converts the referring utterance $\mathcal{U}$ into a symbolic expression $\mathcal{E}$, which encapsulates the spatial relations and category names in $\mathcal{U}$.
Category features, quantifying how well each object matches the category, are derived from the classification results.
Our spatial relation encoders are Python code generated by LLMs. 
Relation features, quantifying the probability of spatial relationships between objects in $\mathcal{E}$, are computed by the relation encoders by explicit geometric calculations.
For example, $f_{\text{near}}^{(i,j)}$ quantifies the probability that the $i$-th object is near the $j$-th object. 
Our executor has a similar design to \cite{hsu2023ns3d}. 
Given the symbolic expression $\mathcal{E}$ and features, An executor uses the $\mathcal{E}$, relation features, and category features to calculate the matching scores between all objects and the referring objects based on the symbolic expression.
Objects with higher matching scores are selected as the candidates. Then \ourmethod employs VLMs to find the target object from the images of these candidates.
Please see Appendix~\ref{appendix:details} for more details.

\subsection{Spatial Relation Encoders}
Sizes and positions of objects in 3D scenes inherently determine spatial relations. For example, the \texttt{near} relation depends on pairwise distances, while \texttt{large} is determined by object volumes. 
In \ourmethod, each spatial relation encoder is a Python class that can compute its associated relation features given the object bounding boxes.

\begin{figure}[ht]
    \centering
    \includegraphics[width=\linewidth]{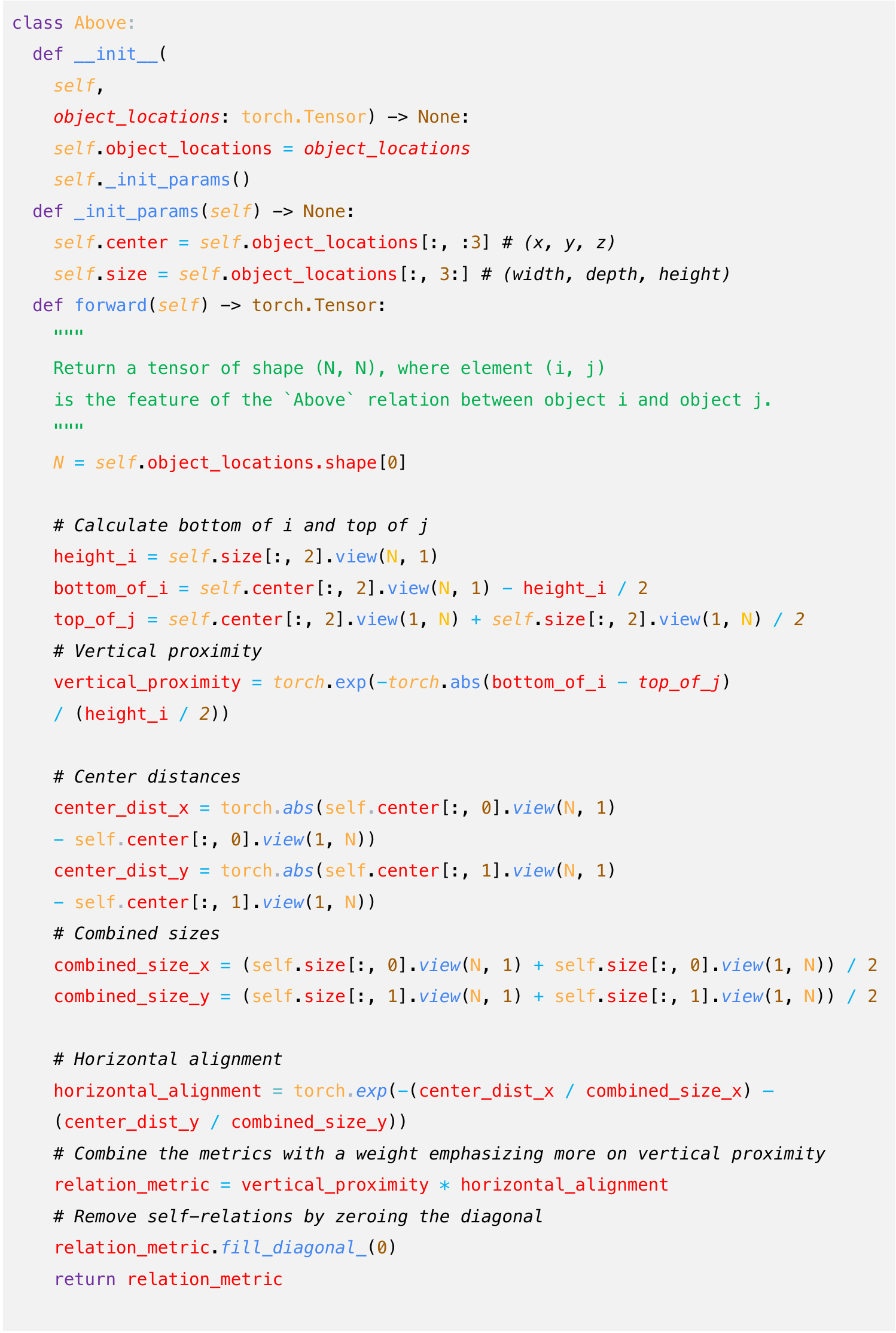}
    \caption{Example of spatial relation encoder for relation ``above''. }
    \label{fig:example-encoder}
\end{figure}

Figure~\ref{fig:example-encoder}
shows the spatial relation encoder of ``above''.
The class is initialized with the object 3D bounding boxes of the scene and provides two key methods: 
\texttt{\_init\_param}, which computes the necessary parameters for feature derivation. For instance, in the ``near'' encoder, it calculates distances between each pair of objects; \texttt{forward}, which performs numerical operations on parameters and returns the relation feature. Specifically, ``above'' encoder computes objects' sizes, horizontal and vertical distances between object pairs to compute the ``above'' feature.

\begin{figure}[h]
    \centering
    \includegraphics[width=\linewidth]{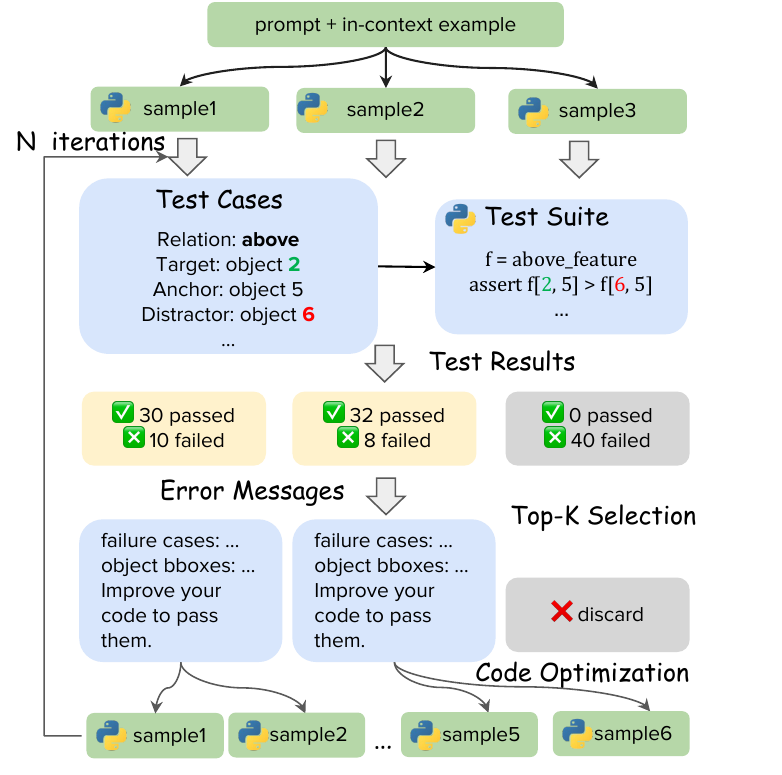}
    % \vspace{-0.3cm}
    \caption{Overview of the generation and optimization process of relation encoders.}
    \label{fig:refinement}
\end{figure}

As illustrated in Figure~\ref{fig:refinement}, the spatial relation encoders come from many optimization iterations.
There are several phases in one iteration: 
(1) retrieving in-context examples based on the semantic similarities of relations (Section~\ref{sec:prompt}); (2) generating multiple codes from LLMs; and (3) testing codes through test suites (Section~\ref{sec:training_data}). When test failures occur, the test suites automatically synthesize error messages that contain failure cases. The codes with highest pass rates and their error messages are then given to LLMs for code optimization (Section~\ref{sec:generation}).

\subsubsection{In Context Example}
\label{sec:prompt}

Adding in-context example into the prompt can improve the response quality from LLMs~\cite{brown2020language}.
To reduce human annotation and provide suitable in-context example for different relation encoders' generation, we retrieve generated codes as in-context example.
For example, relation encoders for ``near'' and ``far'' may both compute pairwise distances but differ only in the numerical processing, so the codes for ``near'' can be used as the in-context example for the generation of ``far''. 
Please see Appendix~\ref{sec:retreiver} for the details.

\subsubsection{Test Suites}
\label{sec:training_data}
To increase the probability of getting high-quality codes, we sample multiple responses from LLMs~\cite{wu2024inference} and design test suites that can evaluate the codes by testing their pass rates on a series of test cases.
Take the relation ``above'' as an example. We collect 37 triplets (less than 100 for most relations) in the format of \texttt{[target object ID], [distractor ID], [anchor object ID]} from the training set, with each triplet serving as a test case.
In relation feature $f_{above}$, if the element $f^{(distractor,anchor)}$ is larger than  $f^{(target,anchor)}$, the test is deemed to have failed, and an error message looks like \texttt{[target bbox] is above [anchor bbox] So feature value of [target bbox] "above" [anchor bbox] should be larger than the feature value of [distractor bbox] "above" [anchor bbox].} is synthesized. An example of such an error message is in 
Section~\ref{sec:prompts}.

\subsubsection{Code Generation and Optimization}
\label{sec:generation}

For any relation, we begin by prompting the LLMs with the task description, the relation name, and the retrieved in-context example~(Section~\ref{sec:prompt}). 
Then we sample $N_{\text{sample}}$ codes from LLMs, where $N_{\text{sample}}$ is a configurable hyperparameter. 
Next, each generated code is tested using the test suites. We select the $top_k$ codes that have the highest pass rates on the test cases and subject them to an optimization phase. 
During the optimization phase, LLMs receive the initial prompt, the code to be optimized, and the error message produced by the test suites. Then LLMs are asked to revise the codes according to failure cases in the error message.
This test and optimization procedure is repeated for up to $N_{\text{iter}}$ iterations. Ultimately, we adopt the code that achieves the highest pass rate across all test cases. 
The detailed optimization and selection algorithm is shown in Algorithm~\ref{alg:code_optimization}.

\subsection{Visual Decision Module}

\label{sec:vlm}

The visual information, like color or shape in utterances, is also essential for accurate grounding, particularly for natural datasets like Nr3D~\cite{achlioptas2020referit3d}.
When two candidate objects share a similar class and spatial position, visual information is required to distinguish between them.

\begin{figure}
    \centering
    \includegraphics[width=\linewidth]{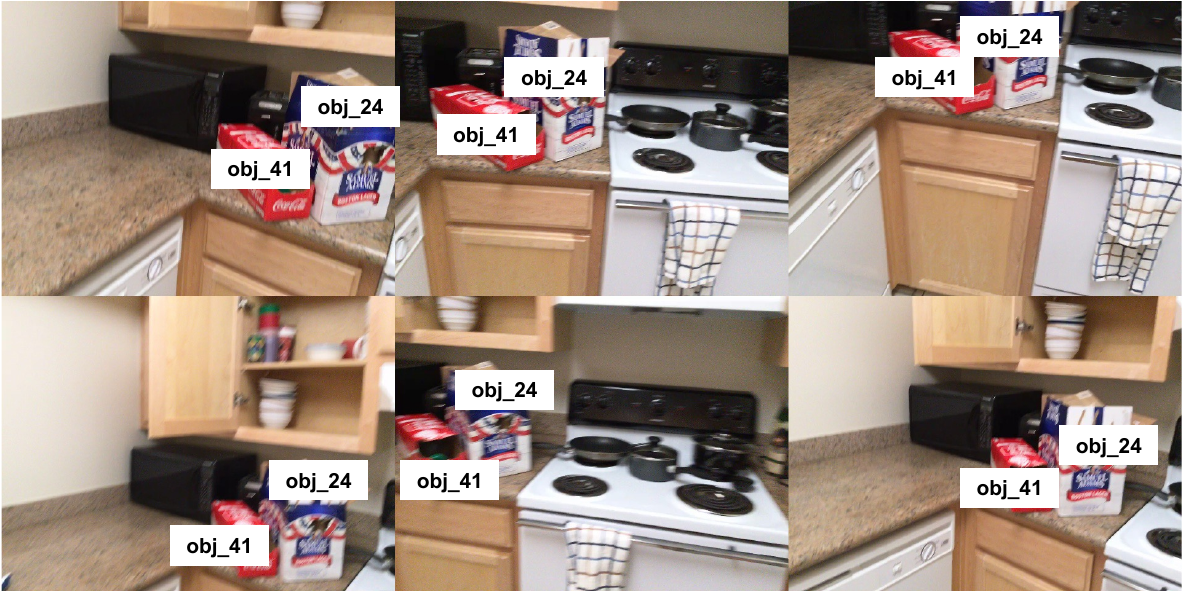}
    \caption{An example of stitched images for VLM prompting. Object ID is annotated on each object's position.
    VLMs can figure out the target ``red'' box from the two candidates and output its ID.
    }
    \label{fig:vlm_example}
\end{figure}

Following VLM-Grounder~\cite{xuvlm}, we incorporate GPT-4o to identify the target object from a set of candidates by utilizing 2D images from ScanNet~\cite{dai2017scannet} as additional context. 
Specifically, we select the top five objects having the highest scores based on the matching scores from the executor and retain those whose logits exceed a chosen threshold as candidates. 
Eight images most relevant to candidate objects from scan images of ScanNet~\cite{dai2017scannet} are selected based on the projected area size of candidate objects. 
They are stitched together in $4 \times 2$ grids and annotated with object IDs.
Finally, we prompt GPT-4o to identify the target object from the stitched images. 
By integrating these visual cues, the VLM decision module effectively disambiguates candidates that appear similar in terms of category and spatial attributes, yielding more accurate grounding results.
An example is in Figure~\ref{fig:vlm_example} (we only show 6 of them for clarity).

\section{Experiments}
\label{sec:exp}
\begin{table*}[h]
\centering
\caption{Performances on Nr3D. 
$\dagger$: For VLM-Grounder~\cite{xuvlm}, we use the results on a 250-sample subset reported in its original paper. 
Results of Transcrib3D in GPT-4o are reported by concurrent SORT3D \cite{sort3d}.
$*$: We re-run ZSSVG3D~\cite{yuan2024visual} in GPT-4o.
}
\begin{center}
\label{tab:nr3d}

\begin{tabular}{lcccccc}

\toprule
\multicolumn{1}{l}{Method}  &\multicolumn{1}{c}{\bf Overall} &\multicolumn{1}{c}{\bf Easy} &\multicolumn{1}{c}{\bf Hard} &\multicolumn{1}{c}{\bf View Dep.} &\multicolumn{1}{c}{\bf View Indep.} 
\\ \hline 
% \texttt{supervised methods} &&&&&\\
\texttt{Supervised} &   & &  & &  \\ 
ViL3DRef~\cite{chen2022language} & \textbf{64.4} & \textbf{70.2} & \textbf{57.4} & \textbf{62.0 }& \textbf{64.5}\\ 
CoT3DRef~\cite{bakr2023cot3dref} &  64.4 & 70.0 & 59.2 & 61.9 & 65.7\\ 
3D-VisTA~\cite{3dvista} & 64.2 & 72.1 & 56.7 & 61.5 & 65.1\\ 
BUTD-DETR~\cite{jain2022bottom} & 54.6 & 60.7 & 48.4 & 46.0 & 58.0\\ 
SAT~\cite{yang2021sat} & 49.2 & 56.3 & 42.4 & 46.9 & 50.4\\ 
\hline
\texttt{Training-free, predicted label} &   & &  & &  \\ 
ZSVG3D$^*$~\cite{yuan2024visual}  & 40.2 & 49.1 & 31.1 & 37.8 & 41.6\\ 
SeeGround~\cite{li2024seeground} & 46.1 & 54.5 & 38.3 &42.3 & 48.2 \\
VLM-Grounder$\dagger$~\cite{xuvlm}  & 48.0 & 55.2 & 39.5 & 45.8 & 49.4\\
\ourmethod w/o VLM & 50.7 & 58.7 & 43.0 & 45.6 & 53.2\\
\ourmethod  & \textbf{52.9} & \textbf{60.7} & \textbf{45.3} & \textbf{49.2} & \textbf{54.7} \\ 
\hline
\texttt{Training-free, ground-truth label} &   & &  & &  \\ 
CSVG~\cite{csvg}  & 59.2 & 59.2 & 44.5 & 53.0 & 46.4\\ 
Transcrib3D~\cite{fang2024transcrib3d}  & 65.6 & - & - & \textbf{63.3} & 66.7\\
\ourmethod w/o VLM &  65.7 & 75.6 & 56.2 & 58.7 & 69.1\\ 
\ourmethod  & \textbf{67.8} & \textbf{76.3} & \textbf{59.6} & 61.6 & \textbf{71.0}\\ 
\bottomrule
% \hline
% NS3D~\cite{hsu2023ns3d} & 52.8 & - & - & - & - \\ 
% \ourmethod w/o VLM & \textbf{60.2}  & - & - & - & - \\
% \bottomrule
\end{tabular}

\end{center}
\end{table*}

\subsection{Experimental Settings}
\paragraph{Dataset}
We mainly conduct experiments on the Nr3D subset of ReferIt3D~\cite{achlioptas2020referit3d} dataset. ReferIt3D has 2 subsets: Nr3D and Sr3D. The Nr3D subset utterances contain human-annotated utterances and the Sr3D contains synthesized ones. 
Based on the number of same-class distractors, the dataset can be categorized into ``easy'' and ``hard'' subsets. The easy subset has a single distractor, and the hard subset has multiple distractors.
The dataset can also be split into ``view dependent'' and ``view independent'' subsets according to the referring utterance. 
Ground truth object bounding boxes are given in the ReferIt3D default evaluation setting. 
Therefore, the metric is an exact match between the predicted bounding box and the target bounding box.
% In ScanRefer, no GT object mask is provided; the evaluation metric is the intersection over union (IoU) value between the predicted bounding box and the GT bounding box. We use the Acc@0.25 metric here. 

% \paragraph{Implementation Details}
% For code optimization (\cref{sec:generation}), we set $N_{sample}$ and $N_{iter}$ to 5, $top_k$ to 3.
% We mainly use \texttt{gpt-4o-2024-08-06} model with a temperature of 1.0 and top\_p of 0.95. 
% For a fair comparison, we use the object classification results from ZSVG3D~\cite{yuan2024visual} for the evaluation of ReferIt3D.
% For the evaluation of ScanRefer, we use the object detection and classification results of ODIN~\cite{jain2024odin}. (Empirically, we find that \ourmethod has similar accuracy when using object detection and Mask3D~\cite{mask3d} segmentation results used in ~\cite{yuan2024visual}.)
% For VLM decision-making, we use the same temperature and top\_p values as VLM-Grounder~\cite{xuvlm}. The thresholds for VLM decision (\cref{sec:vlm}) are 0.9 for Nr3D and 0.1 for ScanRefer.

\paragraph{Baselines}
We compare \ourmethod against both supervised and training-free methods, evaluating accuracy, grounding time, and token cost. The supervised baselines include SAT~\cite{yang2021sat}, BUTD-DETR~\cite{jain2022bottom}, Vil3DRef~\cite{chen2022language}.
% 3D-VisTA~\cite{3dvista}, Chat-Scene~\cite{huang2024chat}.
The training-free approaches include ZSVG3D~\cite{yuan2024visual}, Transcrib3D~\cite{fang2024transcrib3d}, VLM-Grounder~\cite{xuvlm}, CSVG~\cite{csvg} and  
SeeGround~\cite{li2024seeground}.
On the Nr3D dataset, Transcrib3D~\cite{fang2024transcrib3d} and CSVG~\cite{csvg} use ground-truth object labels, providing an advantage over methods which rely on predicted labels; therefore, we compare \ourmethod with them in their specific settings.

\begin{figure*}[!htbp]
    \centering
    \includegraphics[width=\linewidth]{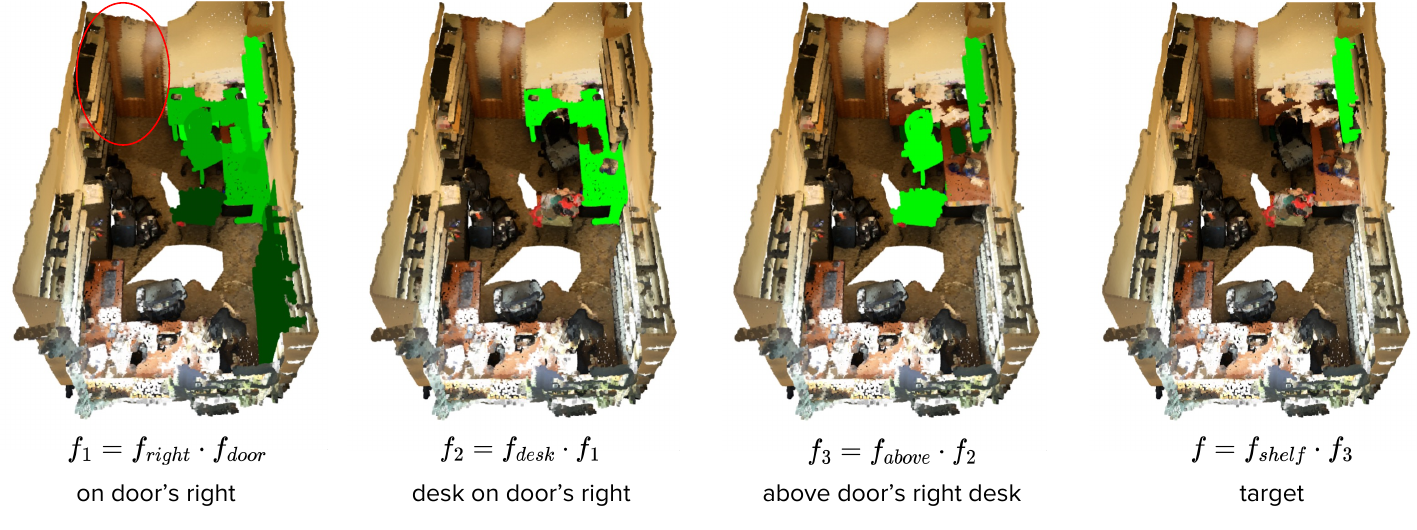}
    \caption{Visualization of the grounding process. Anchor (the door) is marked with \textbf{red circles}. 
    Objects that strongly match the below conditions are highlighted in \textbf{green}, with brighter shades indicating higher matching scores. }
    \label{fig:scene_vis}
\end{figure*}
\subsection{Quantitative Results}
\paragraph{Accuracy}
Table~\ref{tab:nr3d} presents the accuracy comparison on Nr3D.
Compared to other training-free baselines,
\ourmethod achieves higher overall accuracy than both ZSVG3D~\cite{yuan2024visual}, VLM-Grounder~\cite{xuvlm} and SeeGround~\cite{li2024seeground}.
\ourmethod also surpasses one early supervised method, SAT~\cite{yang2021sat} and further narrows the gap in overall performance relative to the supervised method BUTD-DETR~\cite{jain2022bottom}, especially on the view-dependent (VD) subset. 
However, it still lags behind other more recent supervised methods~\cite{chen2022language}, which are trained on large-scale 3D vision-language datasets.
We further evaluate \ourmethod in the experimental settings of~\cite{fang2024transcrib3d, csvg}, in which ground truth object labels are utilized for more accurate category-level object recognition. 
CSVG~\cite{csvg} uses the same spatial functions as ZSVG3D~\cite{yuan2024visual}, resulting a lower accuracy.
Transcrib3D~\cite{fang2024transcrib3d} can produce natural language reasoning processes according to the specific utterance, offering more flexibility.
So that it achieves a close accuracy as \ourmethod.
For SeeGround~\cite{li2024seeground} and CSVG~\cite{csvg}, we use the results reported in the original paper in Table~\ref{tab:nr3d}. 
For a fair comparison, We evaluate \ourmethod and SeeGround~\cite{li2024seeground} using the same VLMs~\cite{qwen2} on Nr3D subset, the overall accuracy are 40.7\% of SeeGround and 48.8\% of \ourmethod.

\begin{table}[t]
\caption{Grounding time and token costs. 
\ourmethod has significant advantage, especially when compared to agent-based methods (VLM-Grounder~\cite{xuvlm} and Transcrib3D~\cite{fang2024transcrib3d}).}
  \centering
  \label{tab:efficiency}
  \begin{tabular}{lcc}
    \toprule
    \multicolumn{1}{l}{Method}  & \multicolumn{1}{c}{\bf Time/s} & \multicolumn{1}{c}{\bf Token}
    \\ \midrule
    ZSVG3D  & 2.4 & 2.5k\\ 
    VLM-Grounder  & 50.3 & 8k\\
    Transcrib3D & 27.0 & 12k\\
    CSVG & 4.0 & 4.0k \\
    SeeGround & 9.0 & 2.6k \\ 
    \ourmethod (w/o VLM) & \textbf{2.1} & \textbf{1.2k} \\
    \ourmethod  & 7.7 (+5.6) & 3.1k (+1.9k)\\ 
    \bottomrule
    % \vspace{-1cm}
  \end{tabular}
\end{table}

\paragraph{Grounding Costs}

Table~\ref{tab:efficiency} compares the average grounding time and token costs of training-free methods on a randomly sampled subset of Nr3D. 

For every referring utterance, Transcrib3D~\cite{fang2024transcrib3d} calls the LLMs for many turns until the target object is found and the context keeps growing, which exhibits significantly higher time and token consumption (27.0s and 50.3k tokens).
In contrast, all codes of \ourmethod are generated before grounding and reused.
So for every referring utterance, \ourmethod calls the LLMs (for parsing) and VLMs for only once.
VLM-Grounder~\cite{xuvlm} inputs all scan images into VLMs, but the executor of \ourmethod can filter out most of the objects so \ourmethod only needs to input one image into VLMs. 
As a result, \ourmethod maintains a large reduction in grounding time and token consumption compared to them.
\ourmethod (without VLMs) and ZSVG3D~\cite{yuan2024visual} only need one LLM call for each referring utterance, so they have similar grounding costs, but \ourmethod demonstrates a significant improvement in accuracy over ZSVG3D~\cite{yuan2024visual}. 
CSVG~\cite{csvg} needs to call LLMs three times for an utterance, causing longer time costs.
SeeGround~\cite{li2024seeground} and \ourmethod call both LLMs and VLMs once for an utterance, thus have a similar time costs.

Above quantitative results underscore the ability of \ourmethod~to balance accuracy and efficiency:  \ourmethod achieves competitive accuracy compared to the most accurate 
training-free methods while offering substantial computational costs.

\paragraph{Code Generation Costs}
The generation and optimization of our relation encoders is a one-time, offline process. The number of optimization iterations varies by the complexity of the spatial relation: \texttt{right} and \texttt{between} require a single iteration, \texttt{left} requires three, while \texttt{above}, \texttt{below}, and \texttt{corner} require up to five. Within each iteration, we query the GPT-4o API up to 25 times for code refinement. Each refinement query, an example of which is provided in Listing.~\ref{feedback}, consumes fewer than 2,000 tokens. Crucially, since the final optimized encoders are reused for all test instances, this one-time development cost is effectively amortized over the entire dataset, rendering its contribution to the average per-instance grounding cost negligible.
\subsection{Qualitative Results}

\begin{figure}
    \centering
    \includegraphics[width=\linewidth]{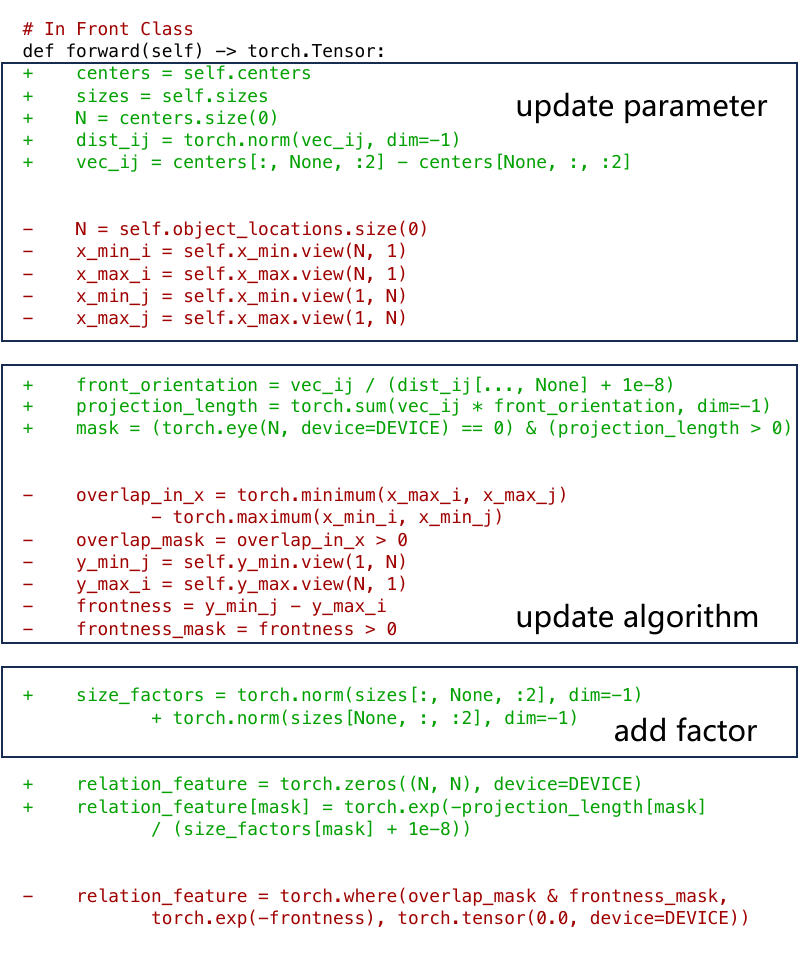}
    \caption{The LLM-based optimization of ``front'' relation encoder.}
    \label{fig:optimize_effect}

\end{figure}
\paragraph{Visualization}

Figure~\ref{fig:scene_vis} visualizes a grounding process of \ourmethod, demonstrating how the final grounding result is constructed through the combination of multiple features of conditions in the referring utterance.
The example referring utterance is ``When facing the door, it’s the shelf above the desk on the right''. 
It can be understood as following four steps in the figure.
First, the feature of \texttt{objects on door's right}, $f_1$, is identified using the category feature ``door'' and the relation feature ``right''.
Next, the feature of \texttt{desk on door's right} is computed by multiplying the category feature ``desk'' and the $f_1$.
Objects \texttt{above on door’s right desk} are identified by relation feature $f_{above}$ and the previous feature; the target ``shelf'' is found by multiplying $f_{shelf}$.
More visualization results can be found in Figure.~\ref{fig:more_vis}.

% \paragraph{Cross Dataset Generalization}
% Although \ourmethod doesn't need training on large-scale 3D datasets, it uses a small part of Referit3D~\cite{achlioptas2020referit3d} to guide the optimization process, where other training-free methods use none of them.
% To validates the generalization of relation encoders, we conduct extra experiments on GRScenes~\cite{grutopia}, a high-quality indoor 3D scene dataset.
% We select 5 scenes and annotate 40 referring utterances.
% For baseline, we randomly select an object with the same category as the target.
% \ourmethod achieves 90.0\% accuracy on the test sets, while the random baseline acieves 15.6\%.
% The results turns out that even if out relation encoders only optimized  by ScanNet and Referit3D language, it can still generalize to different scenes.
% See the appendix for qualitive results.

\paragraph{Cross-Dataset Generalization}
Although \ourmethod requires no pre-training on large-scale 3D datasets, it does exploit a small subset of the ReferIt3D corpus~\cite{achlioptas2020referit3d} during optimization, whereas other training-free approaches use no external data at all.
To probe the generalization ability of our relational encoders, we further evaluate on GRScenes~\cite{grutopia}, a high-quality indoor 3D scene dataset.
We manually annotate 40 referring expressions across five scenes 
and deliberately selected target objects that have many of same-category distractors in the scene.
We adopt a naïve baseline that randomly selects an object from the same semantic category as the target.
\ourmethod attains an overall accuracy of 90.0\%, while the random baseline achieves only 15.6\%.
These results demonstrate that, even when optimized solely with ScanNet~\cite{dai2017scannet} and ReferIt3D~\cite{achlioptas2020referit3d} language, our relational encoders transfer robustly to previously unseen environments.
Qualitative visualizations are provided in Appendix~\ref{sec:cross_data_results}.

\begin{table}[t]
\caption{Accuracy on GRScenes~\cite{grutopia}.}
  \centering
  \label{tab:grscene}
  \begin{tabular}{lccccc}
    \toprule
    \multicolumn{1}{l}{Method}  & \multicolumn{1}{c} {View Dep.}& \multicolumn{1}{c}{View Indep.}   
    \\ \midrule
    \ourmethod   & 87.5\% & 91.7\% \\ 
    Random   &  15.4\% &  15.7\% \\
    \bottomrule
    \vspace{-1cm}
  \end{tabular}
\end{table}

\paragraph{Comparison with Human Annotations}
Eureka~\cite{ma2023eureka} demonstrates that LLMs can surpass human experts in reward-function design.  
For 3D visual grounding, ZSVG3D~\cite{yuan2024visual} relies on manually crafted spatial-relation functions, whereas \ourmethod achieves substantially higher performance.  
Because the two pipelines differ considerably, it is non-trivial to directly reuse their code in \ourmethod.  
To quantify the gap, we evaluate LLM-generated programs and human-written functions on Nr3D~\cite{achlioptas2020referit3d}.  
Substituting the automatically synthesized functions with human-designed ones causes the overall accuracy of \ourmethod to fall sharply to 44.0\%.  
These results highlight the advantages our generated codes over manual annotations.

\paragraph{Optimization Quality}

% By reflecting on failed test cases, LLMs can iteratively optimize relation encoders across multiple dimensions. 
% Figure~\ref{fig:optimize_effect} illustrates this process using the relation``front” as an example, highlighting the difference between the initial LLM-generated implementation (in red) and the optimized version after several optimization steps (in green). 
% Three key improvements are made: First, while the initial implementation relies solely on axis-aligned bounding box coordinates, the optimized version incorporates both distances and directional vectors between object centers. 
% Second, the original code determines ``front” relations based only on X-axis overlap and a numerical comparison along the Y-axis, but the optimized version uses vector projection to assess frontness in arbitrary directions.
% Object size is also introduced as a normalization factor, improving the accuracy and robustness of the relation prediction.
By analyzing failed test cases, LLMs can iteratively refine relation encoders across multiple dimensions. 
Figure~\ref{fig:optimize_effect} illustrates the difference between the initial LLM-generated implementation (in red) and the optimized version (in green) using the relation ``front'' as an example.
The optimized version incorporates both distances and directional vectors between object centers, rather than relying solely on axis-aligned bounding box coordinates.
It also replaces simple X-axis overlap and Y-axis comparison with vector projection, enabling the detection of ``front'' relations in arbitrary directions;.
Additionally, object size is used as a normalization factor, enhancing the accuracy and robustness of the relation prediction.

\paragraph{Robustness to Viewpoints}
We sampled 100 sentences containing explicit viewpoint information (\eg ``facing the...'') from Nr3D. Our analysis revealed that in 98 of these cases, the anchor object of the relation was the same as the object being ``faced'' (e.g., in ``facing the bed, the nightstand on the right'', the bed is the anchor). This suggests that in the vast majority of cases in this dataset, the viewpoint is implicitly defined by the anchor object and does not require separate, explicit modeling. We believe this is plausible, e.g. to judge a relation like ``the left of boxes'', one naturally assumes a viewpoint facing them.

Our extra analysis reveals that our relation encoders have implicitly learned this data-driven prior. For example, our \texttt{left} encoder, when evaluating if object A is to the left of anchor B, operates under the default assumption of a viewpoint facing B. This behavior is not hard-coded, instead it emerges naturally from our test-driven optimization process, aligning with the data's characteristics. This adaptability is further validated by our quantitative results. As shown in Table.~\ref{tab:nr3d}, our method achieves advanced performance on the view-dependent split.

\subsection{Ablation Study}
\label{sec:ablation}
\begin{figure*}[!htbp]
    \centering
    \includegraphics[width=\linewidth]{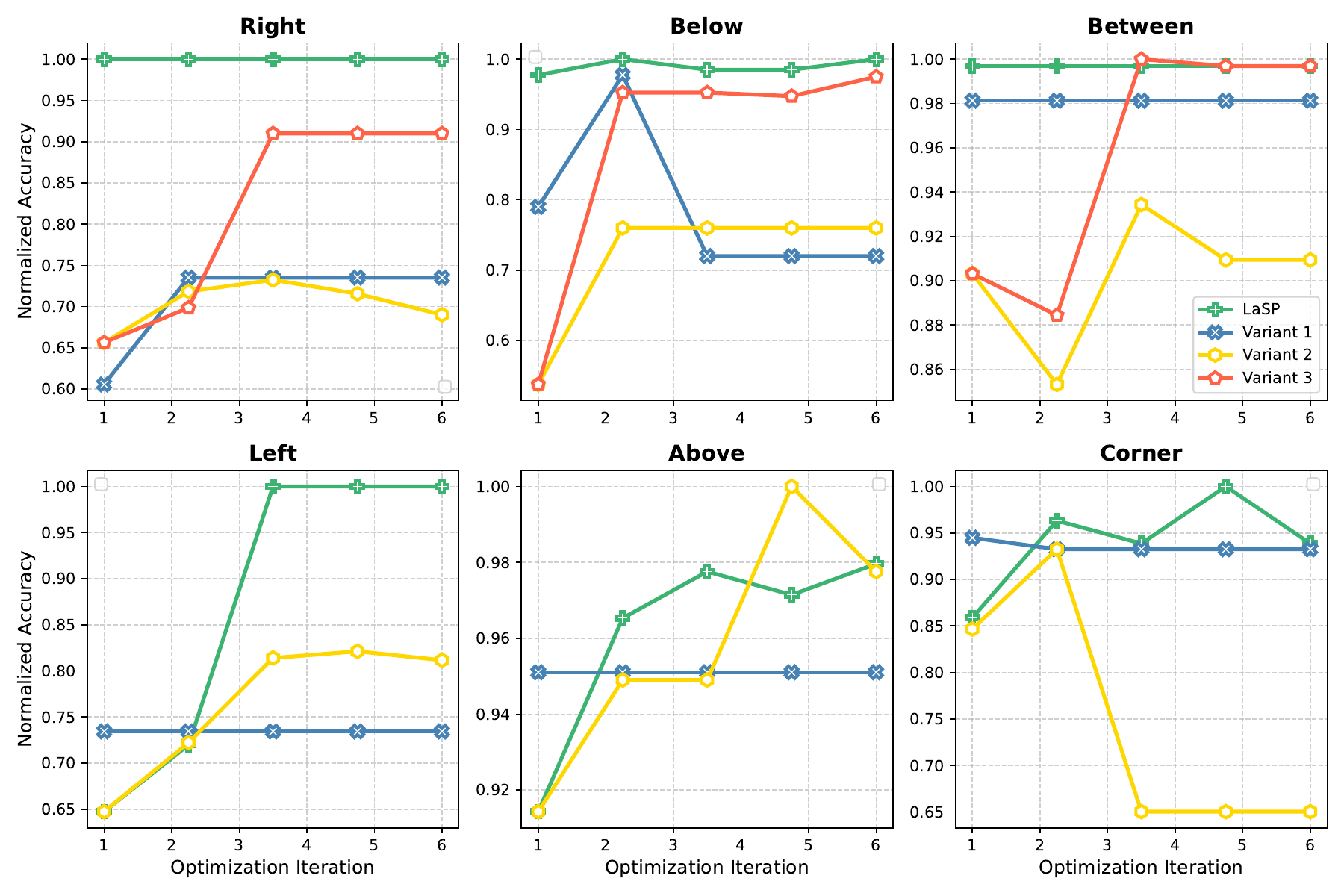}
    \caption{
    \textbf{Ablation study} on different variants. 
    The x-axis stands for the number of optimization iterations.  
    The y-axis stands for the normalized accuracy on corresponding Nr3D subsets. 
}
    \label{fig:ablation}
    % \vspace{-10pt}
\end{figure*}

We conduct ablation studies to investigate the impact of various components during the code generation and optimization processes by evaluating three different variants. 
Variant 1 ablates all three key components: optimization, error messages, and in-context examples. In this variant, we directly prompt LLMs to generate multiple codes and select the one with the highest pass rate on unit tests.
Variant 2 adds optimization processes and ablates the error message by replacing it with a general optimization instruction that doesn't contain any failure case;
variant 3 only ablates in-context examples. 
For relatively simple relations like ``small'', the generated codes can pass all unit tests in the first generation, so there is no following optimization, so we choose to analyze six relations that required multiple optimization iterations. 
For the relations that no in-context example is used (``left'', ``above'', and ``corner''), variant 3 is identical to the full method, so we only report variant 1 and 2 for these relations.
To control for the impact of the initial generation, we use the same responses of the first iteration across variant 2 and 3.

Figure~\ref{fig:ablation} illustrates the results of the ablation study; different variants are represented by lines of different colors. 
The horizontal axis represents the number of iterations.
The vertical axis shows the normalized accuracy on test examples associated with the relation.
The effect of optimization is evident in variant 1: without optimization, LLMs fail to produce accurate relation encoders for most relations, except ``corner'' and ``between''. Variant 2 demonstrates the effect of optimization: by incorporating simple optimization, the accuracies improve on some relations compared to variant 1. 
However, without the detailed error message, LLMs still can't generate accurate encoders for most relations.
The results of variant 3 highlight the effect of error messages:
by using specific failure cases in error messages, LLMs are able to generate more accurate spatial relation encoders for most relations. 
For relations ``right'', ``between'' and ``below'' which use in-context examples, the accuracies of variant 3 are significantly lower than \ourmethod in the first iteration, underscoring the impact of in-context examples.

\section{Conclusion}
\label{sec:conclusion}
% In this paper, we present \ourmethod, a training-free method for 3D visual grounding that uses Python codes to encode spatial relations and an noval pipeline for automatic code generation and optimization. 
% By leveraging rich spatial knowledge from LLMs, \ourmethod eliminates the need for large-scale 3D vision-language datasets.
% For code qualities, we introduce the novel test suites can evaluate LLM-generated codes and provide optimization guidances. 
% Our self-optimization process can iteratively select better codes and  which utilizes test results improve code .

In this paper, we present \ourmethod, a training-free method for 3D visual grounding that uses Python codes to encode spatial relations, along with a automatic generation pipeline.
Leveraging the rich spatial knowledge in LLMs, \ourmethod eliminates the need for large-scale 3D vision–language datasets.
We introduce novel test suites that evaluate LLM-generated codes and guide their optimization.
According to the test results and feedback, 
superior codes are selected and optimized iteratively by LLMs, yielding more accurate spatial relation encoders.
% A self-refinement loop then  selects  variants , 
Experimental results demonstrate that \ourmethod achieves competitive accuracy compared to previous training-free methods while offering promising advantages in time and token costs.
\newpage
\section*{Limitations}
We acknowledge that \ourmethod{} has limitations in the following respects.% \subsection*{Performance Limitations}

\subsection*{Performance Limitations}
Although we have made progress in balancing accuracy and efficiency, there is still a noticeable gap between training-free methods and recent supervised models that jointly learn object detection and spatial reasoning~\cite{3dvista, wu2025rg, arnaud2025locate3drealworldobject}.  
Current training-free approaches—including ours—focus mainly on spatial reasoning while relying on off-the-shelf 3D detectors and object classifiers.  
Designing more accurate, lightweight 3D perception modules tailored for referring tasks therefore remains an important research direction.

\subsection*{System Limitations}

\paragraph{Symbolic Expression.}
Our symbolic representation captures object categories and pairwise spatial relations, but it struggles with ordinal or non-relation phrases such as “second from the left.”  
A possible solution is incorporating order constraints, as in CSVG~\cite{csvg}.

\paragraph{Relation Coverage.}
For simplicity we restrict ourselves to frequent relations (\textit{near}, \textit{above}, \textit{left}, etc.).  
Low-frequency relations are omitted from the current analysis, which may hide weaknesses on those cases.  
Building 3DVG benchmarks with a wider range of challenging relations would enable deeper evaluation.

\paragraph{Scene Representation.}
We simply model a scene as a set of 3D bounding boxes and ignore shapes, orientations, and functional zones (e.g., bathrooms).  
Enriching the scene graph with such information and developing stronger encodings for LLMs and VLMs are promising directions.

\paragraph{Test Suite Design.}
Our test cases are constructed from Nr3D examples that isolate single, atomic spatial relations (e.g., ``on the left of X''). This methodology does not currently account for complex referring expressions that involve nested or compositional relations (e.g., ``the bottle on the table to the right of the sofa''). Extending our framework to jointly optimize multiple encoders is a promising direction to improve generalization to these more complex linguistic structures.

\subsection*{Dependence on Pre-trained 3D Models}
While \ourmethod{} removes the need for large-scale 3D vision–language datasets such as ReferIt3D~\cite{achlioptas2020referit3d} and SceneVerse~\cite{jia2024sceneverse}, it still depends—like most training-free methods~\cite{yuan2024visual, fang2024transcrib3d, li2024seeground, csvg}—on a pre-trained 3D detector and point-cloud classifier.  
VLM-Grounder~\cite{xuvlm} avoids these components by leveraging strong 2D perception models~\cite{Liu2023GroundingDM, Kirillov2023SegmentA}, but its per-utterance detection cannot be reused across multiple queries in the same scene, resulting in high latency.  
Scene-level 3D object discovery based on 2D models~\cite{Gu2023ConceptGraphsO3} may ultimately remove the remaining dependence on 3D training data.

\section*{Ethics Statement}
The human involvement in this study was a small group of volunteer experts who qualitatively annotated some relation encoders. 
All participants were fully briefed on the purpose of the research, provided written informed consent, and were free to withdraw at any time. No demographic or personally identifiable information was collected, stored, or reported.

\section*{Acknowledgement}
We would like to thank Runsen Xu for the helpful discussions.

{\small
\bibliography{11_references}
}

\appendix

\section{Prompts}
\label{sec:prompts}
In this section, we show the prompts we used.
\Cref{prompt:semantic_parsing} is the prompt for GPT-4o to convert referring utterances into symbolic expressions.
\Cref{prompt:initprompt} is an example of a prompt for relation encoder generation, containing the task description and an in-context example.
\Cref{feedback} is an example of error messages. It is synthesized by the test suites and contains failure cases and optimization guidance.
\Cref{prompt:selfoptim} is the code optimization prompt used in the ablation study.
\setcounter{lstlisting}{0}
\begin{lstlisting}[basicstyle=\fontfamily{\ttdefault}\scriptsize, breaklines=true,caption={Prompt for semantic parsing.}, label={prompt:semantic_parsing}]
You are a skilled assistant with expertise in semantic parsing.

## Task Overview
I will provide you with a sentence that describes the location of an object within a scene. Your task is to convert this description into a JSON format that captures the essential details of the object.

### The JSON object should include:
- **"category"**: string, representing the object's category.
- **"relations"**: a list of relationships between the object and other elements in the scene. Each relationship should be represented as a dictionary with the following fields:
    - **"relation_name"**: string, specifying the type of relationship. The relationship can be:
        - *Unary*: choose from ['corner', 'on the floor', 'against wall', 'smaller', 'larger', 'taller', 'lower', 'within'].
        - *Binary*: choose from ['above', 'below', 'beside', 'close', 'far', 'left', 'right', 'front', 'behind', 'across'].
        - *Ternary*: choose from ['between', 'center', 'middle'].
        Only consider **simple** and **general** relations, donot make complex ones like "left of a blue box", "with dark appearance", "facing the window", etc. You should handle these by logical structures.
        If the relationship is not mentioned in the list, you should choose the most appropriate relation above. **Never** create a new relation name!
    - **"objects"**: a list of objects involved in the relationship. Every object in the list should have the same JSON structure. The list structure depends on the relationship type:
        - *Unary*: The list should be empty.
        - *Binary*: The list should contain one object.
        - *Ternary*: The list should contain two objects.
    - **"negative"**: boolean, indicating if the object is explicitly described as not having this relationship. Set this to True if applicable.

## Guidelines:
- First, generate a plan outlining the object's appearance and relationships based on the sentence. Then, use this plan to create the JSON representation.

## Examples:

### Example 1:
**Sentence**: The correct whiteboard is the one on a table.  
**Plan**: "Correct" does not describe appearance. The appearances are "whiteboard" and "table", and the "whiteboard" is on the "table".  
**Parsed JSON**:
```json
{
    "category": "whiteboard",
    "relations": [
        {
            "relation_name": "above",
            "objects": [
                {
                    "category": "table",
                    "relations": []
                }
            ]
        }
    ]
}
```

... 2 more examples.
\end{lstlisting}

\begin{lstlisting}[basicstyle=\fontfamily{\ttdefault}\scriptsize, breaklines=true,caption={Example prompt for relation encoder generation.}, label={prompt:initprompt}]
You are an expert on spatial relation analysis and code generation.

# Introduction to task
Your task is to write a Python class which can be used to compute the metric value of the existence of some spatial relationship between two objects in a 3D scene given their positions and sizes. Higher the metric value, higher the probability of the two objects have that relation. 

In the class, you will be given the positions and sizes of the objects in the scene. The class should have a method `forward` which returns a tensor of shape (N, N), where element (i, j) is the metric value of the relation between object i and object j.

In the 3D scene, x-y plane is the horizontal plane, z-axis is the vertical axis. 

# Introduction to programming environment

Here is an example class for `Left` relation. The class you write should have the same structure as the example class.

```python
class Left:
    # ...
```
Make sure all tensors are placed on `DEVICE`, which has been defined in the environment.
The code output should be formatted as a python code string: "```python ... ```".

# Some helpful tips

(1) You should only use the given variables, and you should not introduce new variables.
(2) The metric value should be sensitive to the input arguments, which means if the arguments change a little, the value should change a lot.
(3) The metric value should be 0 if the two objects don't have that relation, never set negative values!
(4) Never treat an object as its center point, you must consider the size of the bounding box, just like the example code. Never set an threshold to determine the relation. The value of the relation should be continuous and sparse. 
(5) You should imagine that you are at position (0, 0) to determine the relative positions.
(6) Remember you are **in** the scene and look around, not look from the top. So never use the same way as 2D environment.
...

Propose your method first and then generate the code. Think step by step.
Don't use any axis or specific direction as the reference direction or right direction, your method should work for any perspectives.
\end{lstlisting}

\begin{lstlisting}[basicstyle=\fontfamily{\ttdefault}\scriptsize, breaklines=true, caption={Example error message.}, label={feedback}]
We have run your code on some cases. Here are 3 failure cases:

# Case 1.

Metric value of object tensor([ 0.3992, -0.5619,  0.8831,  0.3921,  0.3476,  0.1059], device='mps:0') "above" object tensor([-0.0432, -0.6965,  0.8483,  0.6526,  0.4943,  0.3061], device='mps:0') should be larger than 0. Metric value of object tensor([ 0.3992, -0.5619,  0.8831,  0.3921,  0.3476,  0.1059], device='mps:0') "above" object tensor([-0.0432, -0.6965,  0.8483,  0.6526,  0.4943,  0.3061], device='mps:0') should be higher than the metric value of object tensor([0.5338, 1.1607, 1.1160, 0.2121, 0.3323, 0.8192], device='mps:0') "above" object tensor([-0.0432, -0.6965,  0.8483,  0.6526,  0.4943,  0.3061], device='mps:0'). 

more 2 cases ...

The first three are the center of the object, the last three are the size of the object. x-y is the horizontal plane and z is the vertical axis.
After test, the pass rate of your code is too low. So you MUST check carefully where the problem is. If you can't find the problem, you should
come up with a new algorithm and re-write your code. 

Don't forget the following tips:
(1) You should imagine that you are at position (0, 0, 0) to determine the relative positions.
(2) Remember you are **in** the scene and look around, not look from the top. So never use the same way as 2D environment.
(3) Don't use any of x-axis or y-axis as your perspective, Your method should work for every perspective.
(4) The horizontal plane is x-y plane.

Please carefully analyze each of the failure case and explain why your code failed to pass it. The reason can be incorrect test case might or your code might not be able to handle some specific cases. Please write your analysis for each of the failure cases.

After the analysis of all cases, you should write the improved code based on your analysis. But **never** modify on the class methods and function parameters. 

Some possible improvement ways:
1. Use a new algorithm to calculate the metric value rather than just modifying the existing code.
2. Consider carefully what other factors might be relevant to the spatial relationship between two objects and use them in your calculation.
3. Check the correctness of the input data and the calculation process.
\end{lstlisting}

\begin{lstlisting}[basicstyle=\fontfamily{\ttdefault}\scriptsize, breaklines=true,caption={Prompt for code optimization of variant 2 in the ablation study.}, label={prompt:selfoptim}]
Reflect on the code above, think carefully how to make it better. For example, check if you ignore some factors that may affect the result or use a wrong method.
Then you must re-write the code in the same format. Remeber all the tips!
\end{lstlisting}

\section{Implementation Details}
\label{appendix:details}
\subsection{Semantic Parsing}

A semantic parser converts the referring utterance $\mathcal{U}$ into a JSON-based symbolic expression $\mathcal{E}$, which encapsulates the spatial relations and category names in $\mathcal{U}$. The symbolic expressions have the following elements:

\begin{itemize}
    \item \textbf{Category}: A string indicating the category of the target object referenced in $\mathcal{U}$.
    \item \textbf{Relations}: A list specifying spatial constraints relative to the target object. Each entry in this list contains:
    \begin{itemize}
        \item \textbf{relation\_name}: A string identifying the spatial relation in $\mathcal{U}$ (e.g., ``near,'' ``above'').
        \item \textbf{anchors}: A list of objects that share the given spatial relation with the target object. Each object is represented as its own JSON entity.
        \item \textbf{negative}: A boolean value which, if set to \textit{true}, denotes that the target object \textbf{should not} exhibit the specified spatial relation.
    \end{itemize}
\end{itemize}

For example, the utterance ``chair near the table'' can be represented as: 
\begin{verbatim}
{"category": "chair", "relations":
[{"relation_name": "near", 
"objects": [{"category": "table"}]}]}
\end{verbatim}

Human-annotated natural language expressions exhibit diverse descriptions of relations, leading to a long-tail distribution of \textbf{relation\_name} in parsed expressions. 
To mitigate this, we define a set of common relation names and prompt LLM to select from them for $\mathcal{E}$ instead of using the original terms from $\mathcal{U}$. 
Based on the number of associated objects, the relations can be categorized into \texttt{unary}, \texttt{binary}, and \texttt{ternary}~\cite{feng2024naturally}. 
For simplicity, attributes that describe properties of a single object, such as ``large'' or ``at the corner'' are treated as special types of unary relations.
The classifications are in Table~\ref{tab:relation_cls}.

\begin{table}[h!]
\centering
\caption{Classification of all relations.}
\label{tab:relation_cls}
\begin{tabular}{@{}l l@{}} 
\toprule
Classification & Relations \\
\midrule
unary & large, small, high, low, on the floor, \\
& against the wall, at the corner \\
\midrule
binary & near, far, above, below, \\ 
 & left, right, front, behind \\
\midrule
ternary & between \\
\bottomrule
\end{tabular}
\end{table}

\subsection{Features}
\paragraph{Category Features}
The category features are the matching scores between the objects in the scene and object categories.
\cite{yuan2024visual} provides the predicted category for each object. For the category feature $f_{\text{category}} \in \mathbb{R}^{N}$ ($N$ is the number of objects), we compute the cosine similarity $sim \in \mathbb{R}^{N}$ between the \texttt{category} and the predicted labels using CLIP \citep{radford2021learning}.
Subsequently, we define the category feature as:
$$
f_{\text{category}} = softmax(100 \cdot sim)
$$

\paragraph{Relation Features}
Relation features, quantifying the probability of spatial relationships between objects in $\mathcal{E}$, are computed by the code-based relation encoders. 
For unary relations, the relation feature $f_{\text{unary}} \in \mathbb{R}^{N}$ ($N$ is the number of objects). The features of the binary relation $f_{\text{binary}} \in \mathbb{R}^{N \times N}$ represent the likelihood that there are binary relations between all possible pairs of objects. 
For example, $f_{\text{near}}^{(i,j)}$ quantifies the probability that the $i$-th object is near the $j$-th object. 
Ternary features follow an analogous pattern for relations involving three objects.

We use the object bounding boxes in the scene to initialize the relation encoders and then call the \texttt{forward()} function to compute the corresponding relation feature, \texttt{f\_rel}. These relation features are also cached in a dictionary $R$ for efficient reuse.

\subsection{Executor}
Our executor has a similar design to \cite{hsu2023ns3d}. 
Given the symbolic expression $\mathcal{E}$ and features, the executor computes the matching score between objects and the referring utterance $\mathcal{U}$.
For each relation in \texttt{relations} field of $\mathcal{E}$, the corresponding relation feature $f_{\text{relation}}$ is multiplied with category feature s $f_{\text{category}}$ of its related objects, yielding intermediate features $\{f_i \in \mathbb{R}^{N}\}_{i=1}^K$ (where $K$ is the number of relations). 
Finally, all intermediate features and $f_{\text{category}}$ are aggregated via the element-wise product to compute the final matching scores between objects and the referring utterance. 
See Algorithm~\ref{alg:execute} for more details.

The detailed execution algorithm is presented in Algorithm~\ref{alg:execute}, utilizing the precomputed category features and relation features. The \texttt{Execute} function runs recursively to compute the $\texttt{matching\_score} \in \mathbb{R}^{N}$ ($N$ is the number of objects).
\begin{algorithm}[ht!]
\DontPrintSemicolon
\SetAlgoVlined
\caption{Executor}
\label{alg:execute}
\SetKwInOut{Input}{\textbf{Require}}
\SetKwInOut{Output}{\textbf{Output}}
\Input{symbolic expression $E$, category features $C$, relation features $R$}

\Output{\texttt{matching\_score}}
\setcounter{AlgoLine}{0}

% Initialize the category feature
$\texttt{f\_category} \gets C[E[\texttt{"category"}]]$ 

% Initialize an empty set for logits
$\texttt{matching\_score} \gets \texttt{f\_category}$ 

% Iterate over each relation in the symbolic expression
\ForEach{$\texttt{item\_rel} \in E[\texttt{"relations"}]$}{
    % Extract relation name and corresponding features
    $\texttt{n\_rel} \gets \texttt{item\_rel}[\texttt{"name"}]$ \;
    
    $\texttt{f\_rel} \gets R[\texttt{n\_rel}]$ \;
    
    % Extract anchors from the relation item
    $\texttt{anchors} \gets \texttt{item\_rel}[\texttt{"anchors"}]$ \;

    % Check the category of the relation name
    \If{$\texttt{n\_rel} \in \texttt{Unary\_Relations}$}{
        % Unary relation: relation_feature is (N,)
        
        $\texttt{f} \gets \texttt{f\_rel}$ \;
    }
    
    \ElseIf{$\texttt{n\_rel} \in \texttt{Binary\_Relations}$}{
        % Binary relation: relation_feature is (N, N)
        % anchors length is 1
        $\texttt{a} \gets \texttt{Execute}(\texttt{anchors}[0])$ \;
        
        $\texttt{f} \gets \texttt{f\_rel} \cdot \texttt{a}$ \; % Matrix-vector multiplication
    }
    \ElseIf{$\texttt{n\_rel} \in \texttt{Ternary\_Relations}$}{
        $\texttt{a\_1} \gets \texttt{Execute}(\texttt{anchors}[0])$ \;
        
        $\texttt{a\_2} \gets \texttt{Execute}(\texttt{anchors}[1])$ \;

        % Einsum notation for tensor computation
        $\texttt{pattern} \gets \texttt{"ijk,j,k}\to\texttt{i"}$ \;
        
        $\texttt{f} \gets \texttt{einsum}(\texttt{pattern}, \texttt{f\_rel}, \texttt{a\_1}, \texttt{a\_2})$ \;
    }
    
    $\texttt{f} \gets \texttt{softmax}(\texttt{f})$ \;

    \If{$E[\texttt{"negative"}]$}{
        % Unary relation: relation_feature is (N,)
        
        $\texttt{f} \gets \texttt{max}(\texttt{f}) - \texttt{f}$ \;
    }
    % Append concept-related matching\_score or other computations (optional)
    $\texttt{matching\_score} \gets \texttt{matching\_score} \cdot \texttt{f}$ \;
}

\Output{\texttt{matching\_score}}
\end{algorithm}

\subsection{Code Generation and Optimization}

Detailed algorithm of code generation and optimization is shown in Algorithm~\ref{alg:code_optimization}.
\begin{algorithm}[ht!]
\caption{Code Generation and Optimization}
\label{alg:code_optimization}
\DontPrintSemicolon
\SetAlgoVlined
\SetKwInOut{Input}{\textbf{Require}}
\SetKwInOut{Output}{\textbf{Output}}
\SetKwInOut{Hyperparameters}{\textbf{Hyperparameters}}

\Input{relation name $R$, relation name $G$, code library $L$, test cases $C$, LLM $\texttt{LLM}$, test suites $T$, initial prompt $\texttt{prompt}$}
\Output{\texttt{best\_code}}

\Hyperparameters{search iteration $N$, sample number $M$, optimizing example number $top_k$}

\BlankLine
\setcounter{AlgoLine}{0}

% --------------------------------------------------------------
% Step 1: Retrieve in-context example and initialize prompt
% --------------------------------------------------------------
$\texttt{example} \leftarrow \texttt{retrieve}(G, R)$ 

$\texttt{init\_prompt} \leftarrow \texttt{prompt} + \texttt{example}$

% --------------------------------------------------------------
% Step 2: Sample initial candidates from the LLM
% --------------------------------------------------------------
$F_1, \dots, F_M \leftarrow \texttt{LLM}(R, \texttt{init\_prompt})$

% --------------------------------------------------------------
% Step 3: Evaluate all candidate codes
% --------------------------------------------------------------
\For{$j \leftarrow 1 \dots M$}{
    $acc_j, err_j \leftarrow T(F_j)$  
    \tcp*[r]{Test each code.}
}

$\texttt{max\_acc} \leftarrow \max \bigl(\{acc_1, \dots, acc_M\}\bigr)$ 

$\texttt{best\_code} \leftarrow F_{\arg\max(\{acc_1, \dots, acc_M\})}$

% --------------------------------------------------------------
% Step 4: Select top-k candidates for refinement
% --------------------------------------------------------------
$\texttt{TopK} \leftarrow \text{SelectTopK}\bigl(\{(F_j, acc_j)\}_{j=1}^M, K\bigr)$

% --------------------------------------------------------------
% Step 5: Iterative refinement of top-k candidates
% --------------------------------------------------------------
\For{$i \leftarrow 2 \dots N$}{
    
    $\texttt{results} \leftarrow []$
    
    \For{$j \leftarrow 1 \dots K$}{
        $(F_{\text{old}}, err_{\text{old}}) \leftarrow \texttt{TopK}[j]$
        
        % Combine code-to-refine with error info
        $\texttt{prompt}_{\text{ref}} \leftarrow \texttt{init\_prompt} + F_{\text{old}} + err_{\text{old}}$
        
        % Sample M new candidates using refined prompt
        $F_1, \dots, F_M \leftarrow \texttt{LLM}(R, \texttt{prompt}_{\text{ref}})$
        
        \For{$k \leftarrow 1 \dots M$}{
            $\texttt{results}.\texttt{append}(F_k)$
        }
    }
    
    % ----------------------------------------------------------
    % Step 6: Evaluate newly refined candidates
    % ----------------------------------------------------------
    $\texttt{eval\_results} \leftarrow []$
    
    \ForEach{$F_k \in \texttt{results}$}{
        $acc_k, err_k \leftarrow T(F_k)$
        
        \If{$acc_k = 1$}{
            \Return $F_k$ 
        }
        
        \If{$acc_k > \texttt{max\_acc}$}{
            $\texttt{max\_acc} \leftarrow acc_k$

            $\texttt{best\_code} \leftarrow F_k$
        }
        
        $\texttt{eval\_results}.\texttt{append}\bigl((F_k, acc_k, err_k)\bigr)$
    }
    
    % ----------------------------------------------------------
    % Step 7: Update TopK for the next iteration
    % ----------------------------------------------------------
    $\texttt{TopK} \leftarrow \text{SelectTopK}(\texttt{eval\_results}, K)$
}

% --------------------------------------------------------------
% Step 8: Store and output the best code
% --------------------------------------------------------------
$L \leftarrow L \cup \{\texttt{best\_code}\}$ \\
\Return{\texttt{best\_code}}
\end{algorithm}

\subsection{In Context Example Selection}
\label{sec:retreiver}
\begin{figure}
    \centering
    \includegraphics[width=\linewidth]{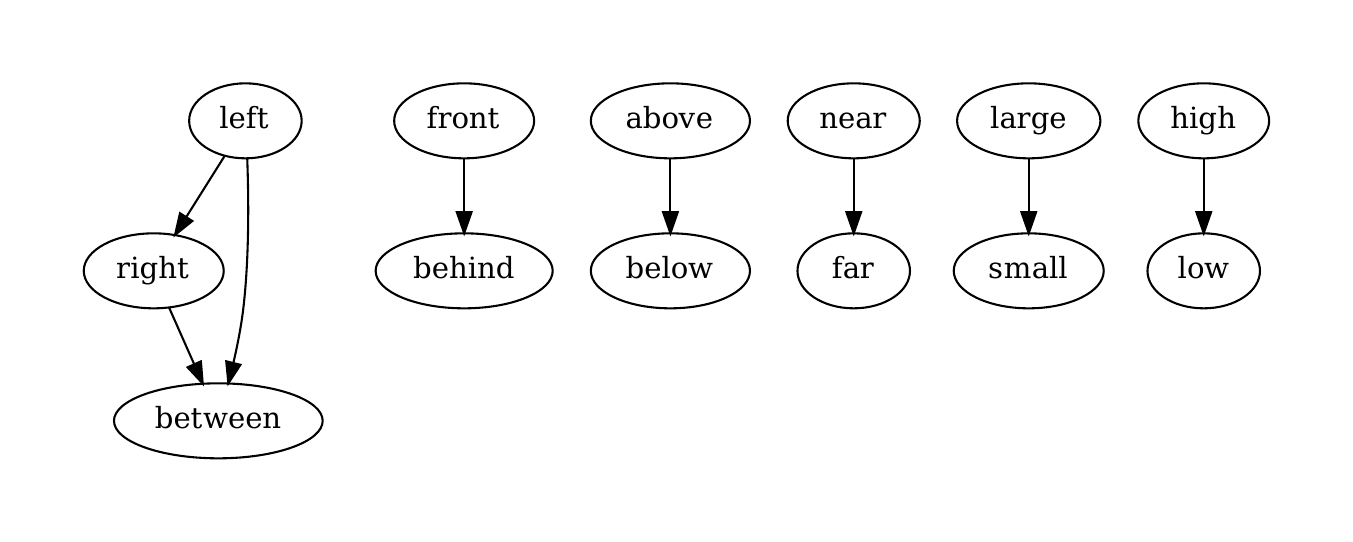}
    \caption{The graph representation for in context example selection.}
    \label{fig:dag}
\end{figure}

The selection of in-context examples is based on relevance. We represent the selection in a graph Figure~\ref{fig:dag}, where an edge from node A to node B means that the encoder for relation A is used as an in-context example when generating for relation B.

\subsection{Hyperparameters and Hardware}
For code optimization (Section~\ref{sec:generation}), we set $N_{sample}$ and $N_{iter}$ to 5, $top_k$ to 3.
We mainly use \texttt{gpt-4o-2024-08-06} model with a temperature of 1.0 and top\_p of 0.95. 
For a fair comparison, we use the object classification results from ZSVG3D~\cite{yuan2024visual} for the evaluation of ReferIt3D.
For VLM decision-making, we use the same temperature and top\_p values as VLM-Grounder~\cite{xuvlm}. 
The thresholds for VLM decision (Section~\ref{sec:vlm}) are 0.9 for Nr3D.
We conduct all experiments on a single NVIDIA GeForce RTX 4090 GPU.

\section{Additional Quantitative Results}
\label{appendix:sr3d_results}
\paragraph{NS3D}
We show evaluation results in NS3D\citep{hsu2023ns3d} in Table~\ref{tab:nr3d}
.
NS3D can only learn concepts (\textit{e.g.} relation name, category name) from the training set and its parsing results of Nr3D contain more than 5,000 concepts, resulting in a long-tailed problem. So it selects a subset containing 1,041 test examples, which only contains the same concepts as Sr3D, the dataset it is mainly trained on. On the NS3D subset, \ourmethod achieves 60.2\% accuracy, NS3D\citep{hsu2023ns3d} have a accuracy of 52.7\%,
which shows the advantage of \ourmethod for processing natural grounding tasks.

\paragraph{Sr3D}

We show evaluation results on Sr3D, a subset of ReferIt3D~\cite{achlioptas2020referit3d} in Table~\ref{tab:sr3d}.
If using predicted object labels, \ourmethod has close accuracy to NS3D~\cite{hsu2023ns3d}.
Even not using training data of Sr3D, \ourmethod still achieves comparable performance with NS3D~\cite{hsu2023ns3d} on both settings (w/ and w/o GT labels).
If using GT object labels, the accuracy of our method (w/o VLM) on Sr3D is 95.3\%, and the performance of NS3D and \cite{fang2024transcrib3d} are 96.9\% and 98.4\%. So \textbf{we believe that the bottleneck of Sr3D performance is object detection and classification rather than spatial relation understanding} because its relation annotation is synthesized by relatively simple functions. 
So we mainly focus on natural benchmarks (Nr3D) which have complex and real spatial relations. 

\begin{table}[t]
\centering
\caption{Performance on Sr3D.}
\begin{center}
\label{tab:sr3d}
\begin{tabular}{lccc}
\hline
\multicolumn{1}{l}{Method}    &\multicolumn{1}{c}{\bf Sr3D} 
\\ \hline 
BUTD-DETR& 67.0 \\ 
NS3D & 62.7 \\ 
NS3D(w/ GT Object Label) & 96.9  \\ 
Transcrib3D (w/ GT Label) & 98.4 \\
\ourmethod (w/o VLM)& 62.0  \\ 
\ourmethod (w/o VLM, w/ GT Object Label) & 95.1 \\
\hline
\end{tabular}
\end{center}
\end{table}

\begin{figure}
    \centering
    \includegraphics[width=1\linewidth]{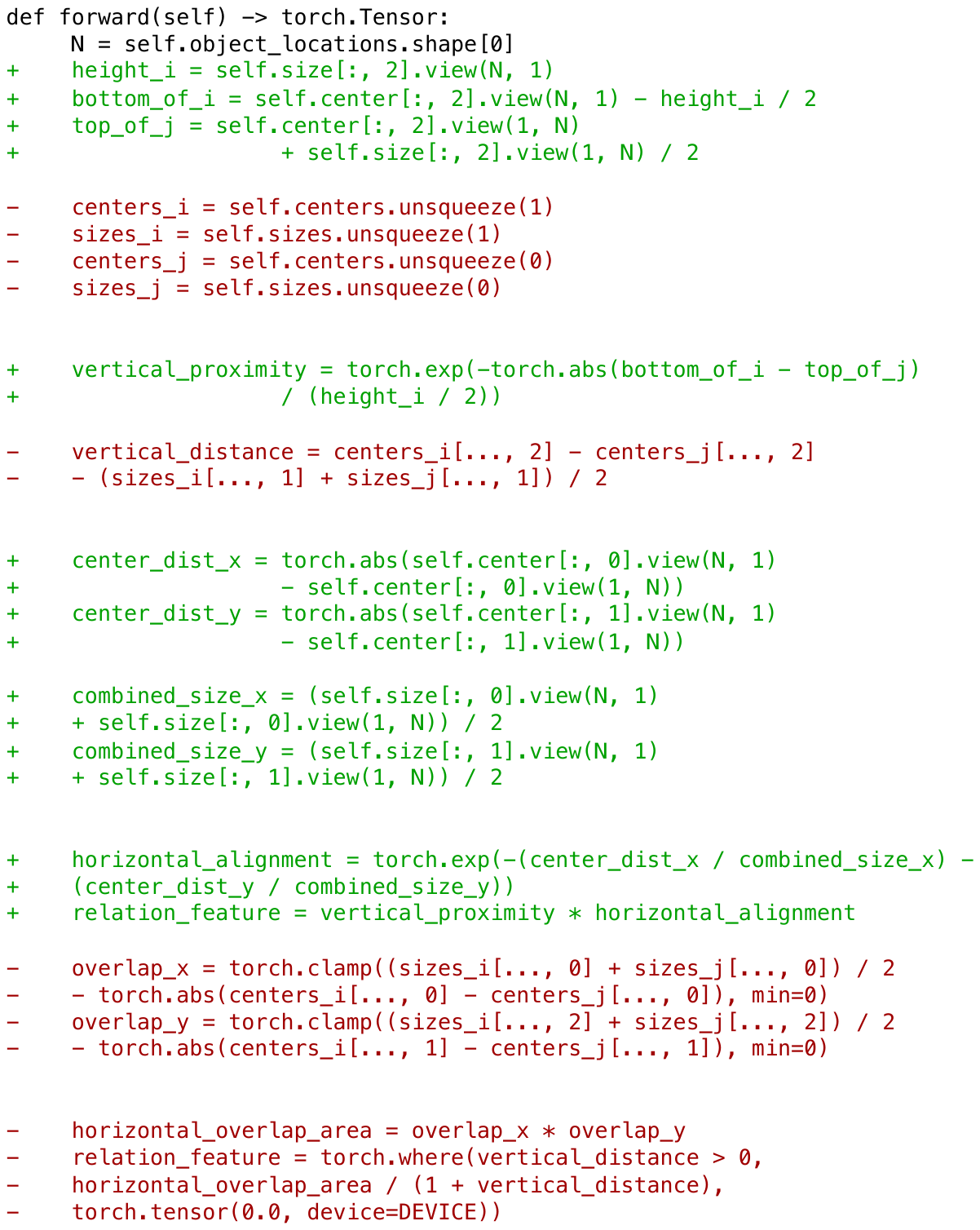}
    \caption{Example of code optimization result on relation encoder of ``above''. }
    \label{fig:above_optim}
\end{figure}

\section{Additional Qualitative Results}
\subsection{Effect of Code Optimization}
\label{app:code_optim}
We show the change between the initial response and the final code after multiple rounds of sampling and iterative refinement in Figure~\ref{fig:above_optim}. 
The initial output only passes 18 test cases out of 37. After 3 iterations of sampling and optimization, we get the the relation encoder as the right one. It passes all 37 test cases.
By transitioning from a strict geometric overlap calculation to a continuous, exponential-based measure for both vertical and horizontal distances, the optimized code now captures nuances in ``above'' relation more robustly. This improved formulation inherently handles scenarios where objects are close but not strictly overlapping, and it provides a more stable gradient for training. Consequently, the updated model passes all test cases by offering a smoother, more differentiable metric that better aligns with real-world spatial relations and passes more test cases.

\subsection{Relation Constraints}
LARC~\cite{feng2024naturally} proposes that certain spatial relations are symmetric, like ``near'' or ``far'', meaning that if object A is ``near'' B, then B should also be ``near'' A. 
Consequently, the features of these relations should be symmetric.
Conversely, other relations are inherently asymmetric, such as ``left'' or ``right''. For these relations, if a feature element is positive (indicating the presence of the spatial relation), its corresponding symmetric element should be negative (indicating the absence of the reverse spatial relation).

\begin{figure*}
    \centering
    \includegraphics[width=\linewidth]{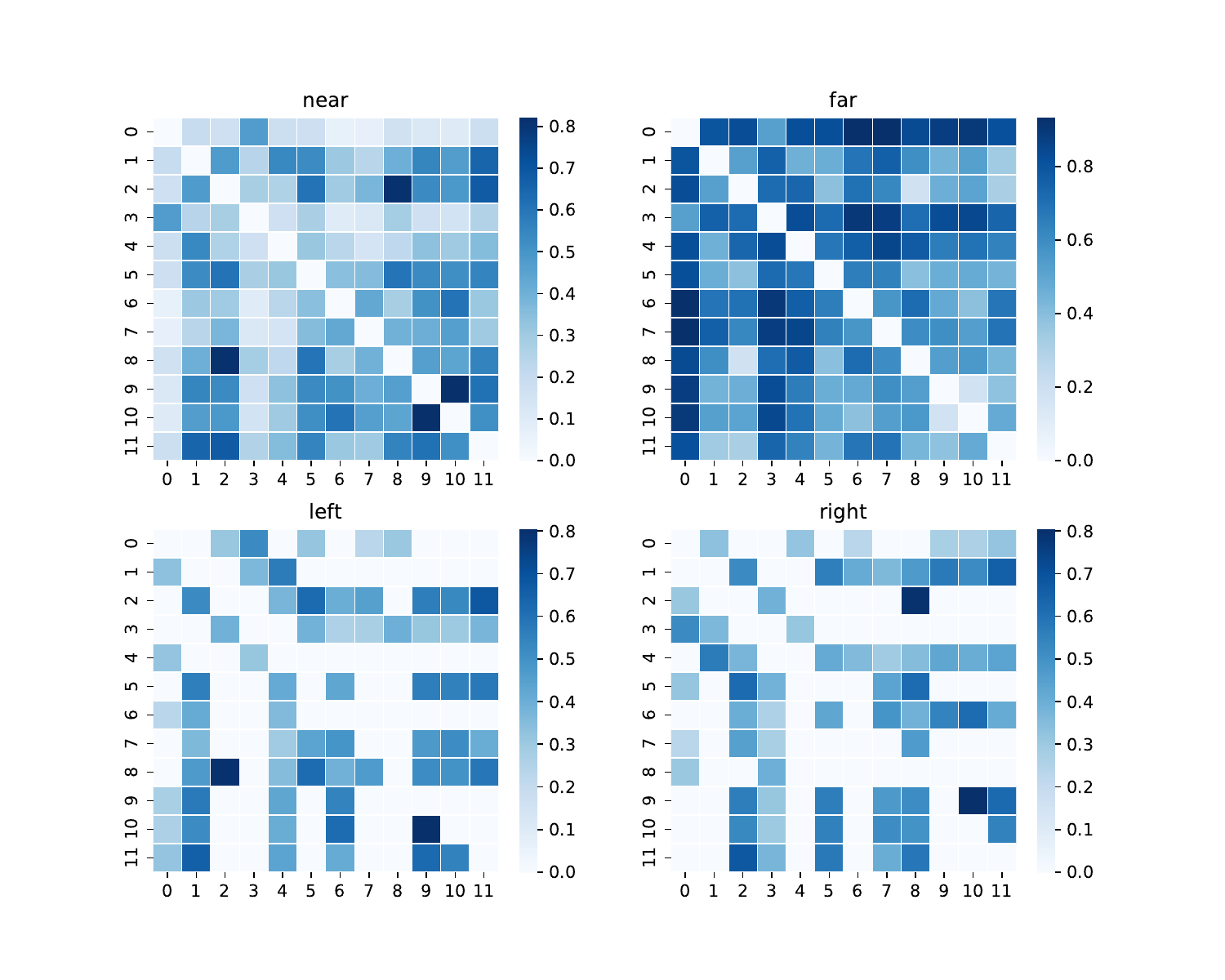}
    \caption{Relation feature examples. The features of ``near'' and ``far'' are symmetric, meaning mutual relationships hold true in both directions. For ``left'' and ``right,'' if an element is positive, its corresponding symmetric element is zero, ensuring asymmetry. Additionally, all diagonal elements are zero, as self-relations are not considered.}
    \label{fig:heatmap}
\end{figure*}

LARC~\cite{feng2024naturally} leverage large language models (LLMs) to annotate these constraints and apply an auxiliary loss to enforce them during training. In contrast, while \ourmethod does not explicitly train or use specialized instructions to create these constraints, we observe that some LLM-generated relation encoders inherently produce relation features that satisfy these constraints. Moreover, for certain relations, these constraints are guaranteed due to the deterministic execution of the code. In Figure~\ref{fig:heatmap}, we present four relation features for a scene:

\begin{itemize} 
    \item Features for ``near'' and ``far'' are guaranteed to be symmetric.
    \item For asymmetric features such as ``left'' and ``right'' if $f_{i, j} > 0$, it is guaranteed that $f_{j, i} = 0$. ''
\end{itemize}

\subsection{Condition Level Accuracy}

% Our parsed symbolic expressions actually contain one or more spatial conditions of the target object. However, there may be some redundant condition in the utterance. 
% For example, if the referring utterance is ``find the monitor on the floor and under desk.''
% , and all monitors ``on the floor'' are ``under the desk'', so one of these two conditions are redundant, which means even if the method can not process one condition of them, it can still give the correct grounding result. 
% So we test \ourmethod on utterances containing single condition for a better understanding of its ability.

% We categorize objects of same class to groups. With in a group, we collect the conditions for each object from parsed expressions. Each condition is a JSON format like 
% \texttt{\{\{''relation'': ..., ''anchors'': [...]\}\}} and can be executed seamlessly to find best matching object. We calculate the average precision and recall for all condition level matches.
% \ourmethod has 67.5\% average precision and 66.9\% average recall.
Our parsed symbolic expressions typically include one or more spatial conditions for the target object. However, some conditions in the referring utterance may be redundant.

For instance, if the referring utterance is ''find the monitor on the floor and under the desk,'' and all monitors ''on the floor'' are also ''under the desk,'' then one of these two conditions is redundant. This means that even if the method fails to process one of the conditions, it can still provide the correct grounding result.
To better understand \ourmethod's capability, we evaluate it on utterances containing a single condition.
We categorize objects of the same class into groups. Within each group, we collect the conditions for each object from the parsed expressions. Each condition is represented in JSON format, such as:
\texttt{{"relation": ..., "anchors": [...]}}.
These conditions are executed seamlessly to identify the best-matching object. We compute the average precision and recall for all condition-level matches. \ourmethod achieves an average precision of 67.5\% and an average recall of 66.9\%.

\subsection{More Visualization Results}
\label{appendix:more_vis_results}

\begin{figure*}
    \centering
    \includegraphics[width=1\linewidth]{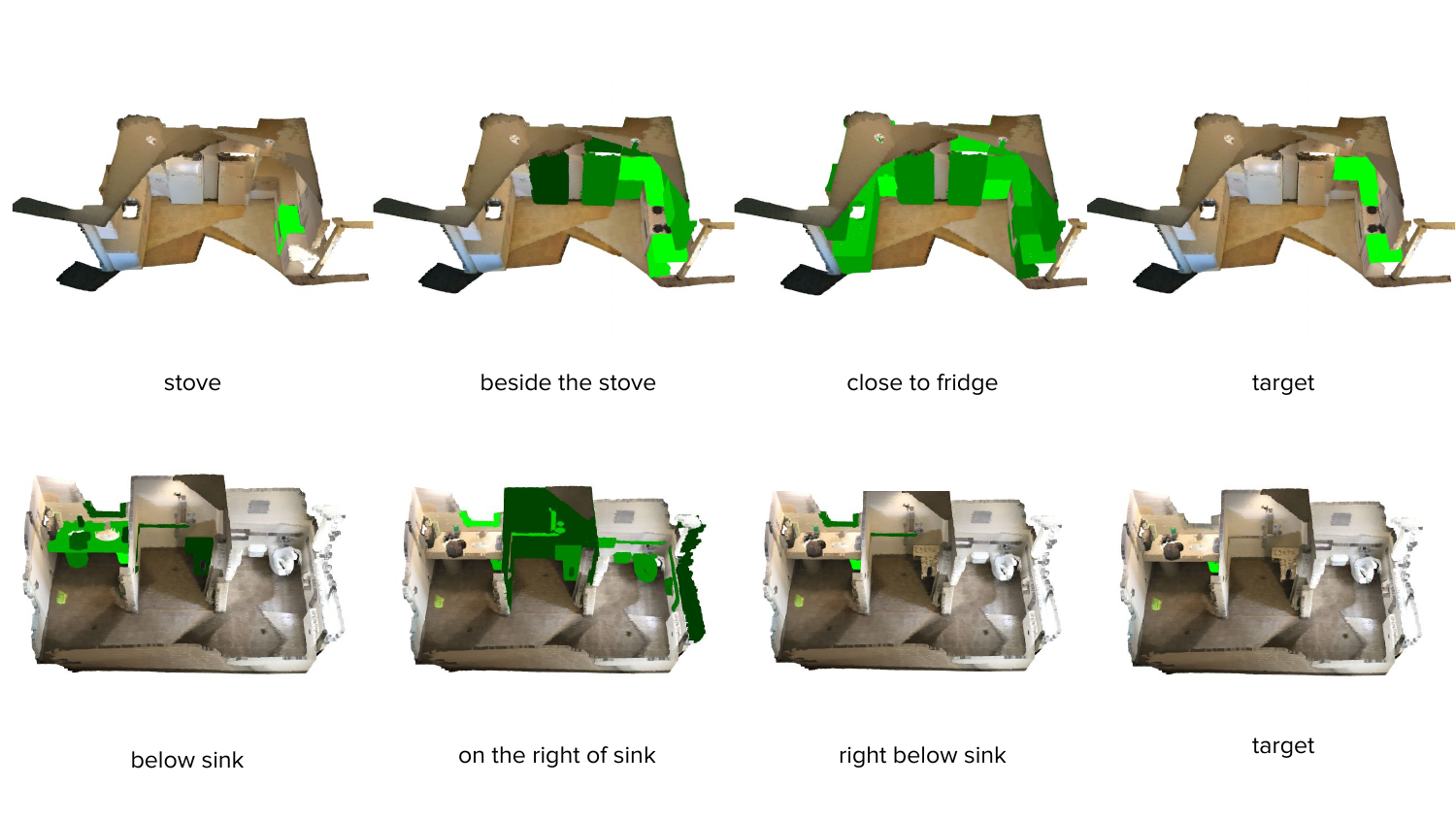}
    \caption{The target objects are: ``Stove next to another stove and close to the fridge'' (top row) and ``Trashcan to the right of and below the sink'' (bottom row).}
    \label{fig:more_vis}
\end{figure*}

We visualize more grounding examples in Figure~\ref{fig:more_vis}.
% The first row shows the process of grounding \texttt{the kitchen cabinet close to fridge and beside the stove.}
% The second row is the grounding process of \texttt{trash can on right below the sink.}
The first row illustrates the grounding process for \texttt{the kitchen cabinet close to the fridge and beside the stove}. In the process, the stove, objects beside the stove, and the objects near the fridge are sequentially grounded, culminating in the target kitchen cabinet highlighted in green.
The second row shows the grounding process for \texttt{right trash can below the sink}. Starting with the objects below the sink, followed by the objects on the right of the sink, and finally combining these conditions to highlight the target trash can in green.

\subsection{Effect of Unit tests}
% To demonstrate the effect of filtering generated code by its accuracy on training cases, we choose 6 relations and plot the pass rate on training cases on the x-axis and the number of passed examples in all relative test examples. For some easy relations like ``near'' or ``far'', GPT-4o can pass all the tests at once, so we only show the cases having multiple refinement steps.

% The result is shown in~\{fig:train_acc_test_acc}. For 5 of 6 relations (except relation \texttt{behind}), the code having the highest performance on the training cases can have the top-tier performance on the test set. As for relation \texttt{behind}, using the best code on training cases causes about 15 cases loss on the test case compared to using about 70 percent accurate code. But it's still better than using most of the codes whose accuracy is less than 0.5. This might be caused by the bias of training data collection. But in general, choosing codes according to performance on the training set is useful for overall performance on the test set.
% \subsection{Test case numbers}
% We list the number of test cases we used for each relation:
To demonstrate the impact of filtering generated code based on its accuracy on training cases, we selected six relations and plotted their performance. The x-axis represents the pass rate on training cases, while the y-axis shows the number of passed examples in all relevant test cases.

For straightforward relations such as ``near'' or ``far'', GPT-4o can pass all unit tests on the first attempt, so we focus on cases requiring multiple refinement steps.

\begin{figure*}
    \centering
    \includegraphics[width=\linewidth]{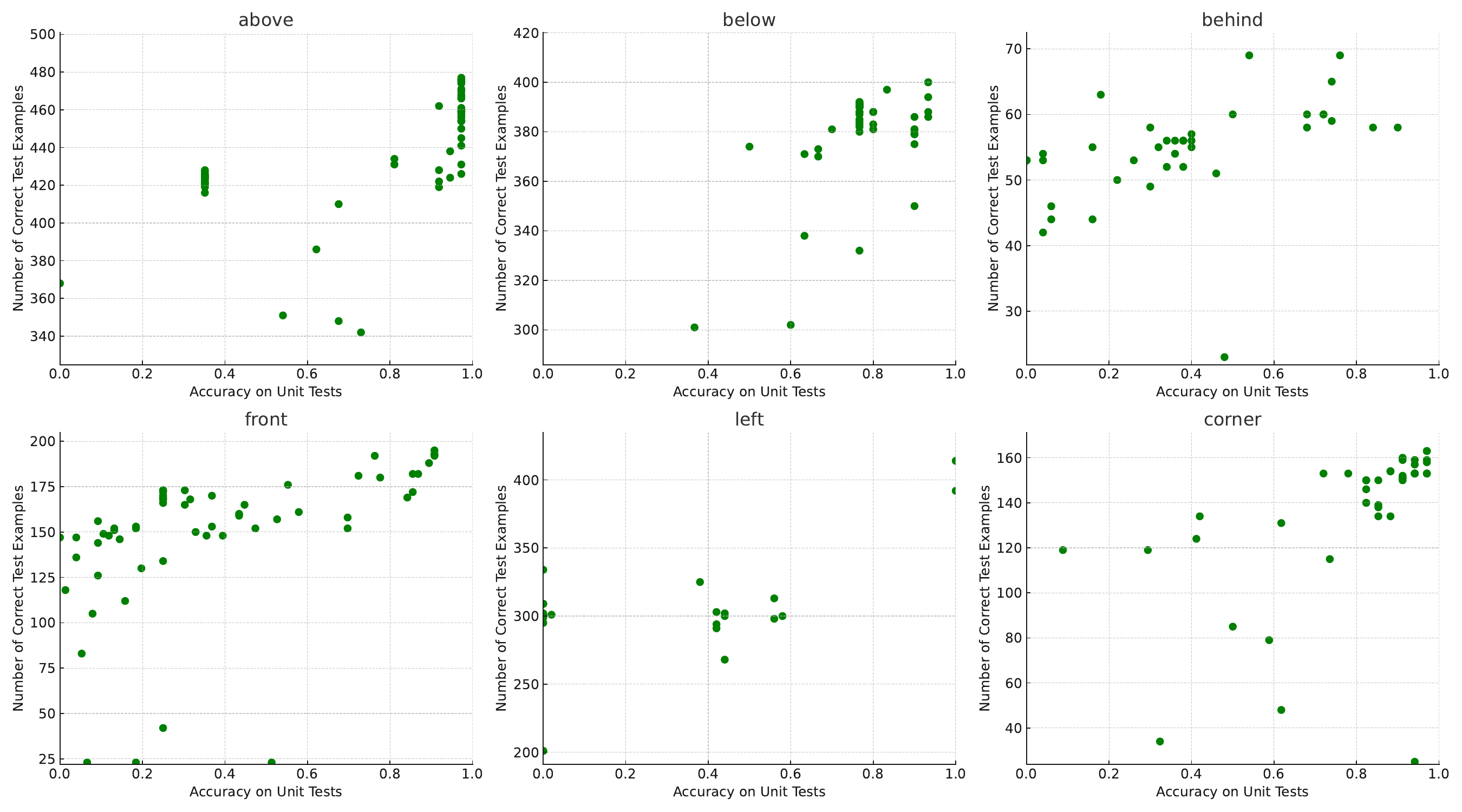}
    \caption{Corresponding relation between the unit test pass rate and number of correct examples on test set.}
    \label{fig:train_acc_test_acc}
\end{figure*}

The results, shown in Figure~\ref{fig:train_acc_test_acc}, indicate that for five out of six relations (excluding \texttt{behind}), the code with the highest pass rate on training cases achieves top-tier performance on the test set. However, for the \texttt{behind} relation, using the best-performing code on the training cases results in about 15 fewer passed test cases compared to using code with approximately 70\% accuracy. Despite this, it still outperforms code with accuracy below 0.5.

This discrepancy for \texttt{behind} may stem from biases in the training data collection process. Overall, selecting code based on its performance on the training set is effective for achieving strong test set performance.

\subsection{Cross Dataset Results}
\label{sec:cross_data_results}
\begin{figure*}[t]
  \includegraphics[width=0.48\linewidth]{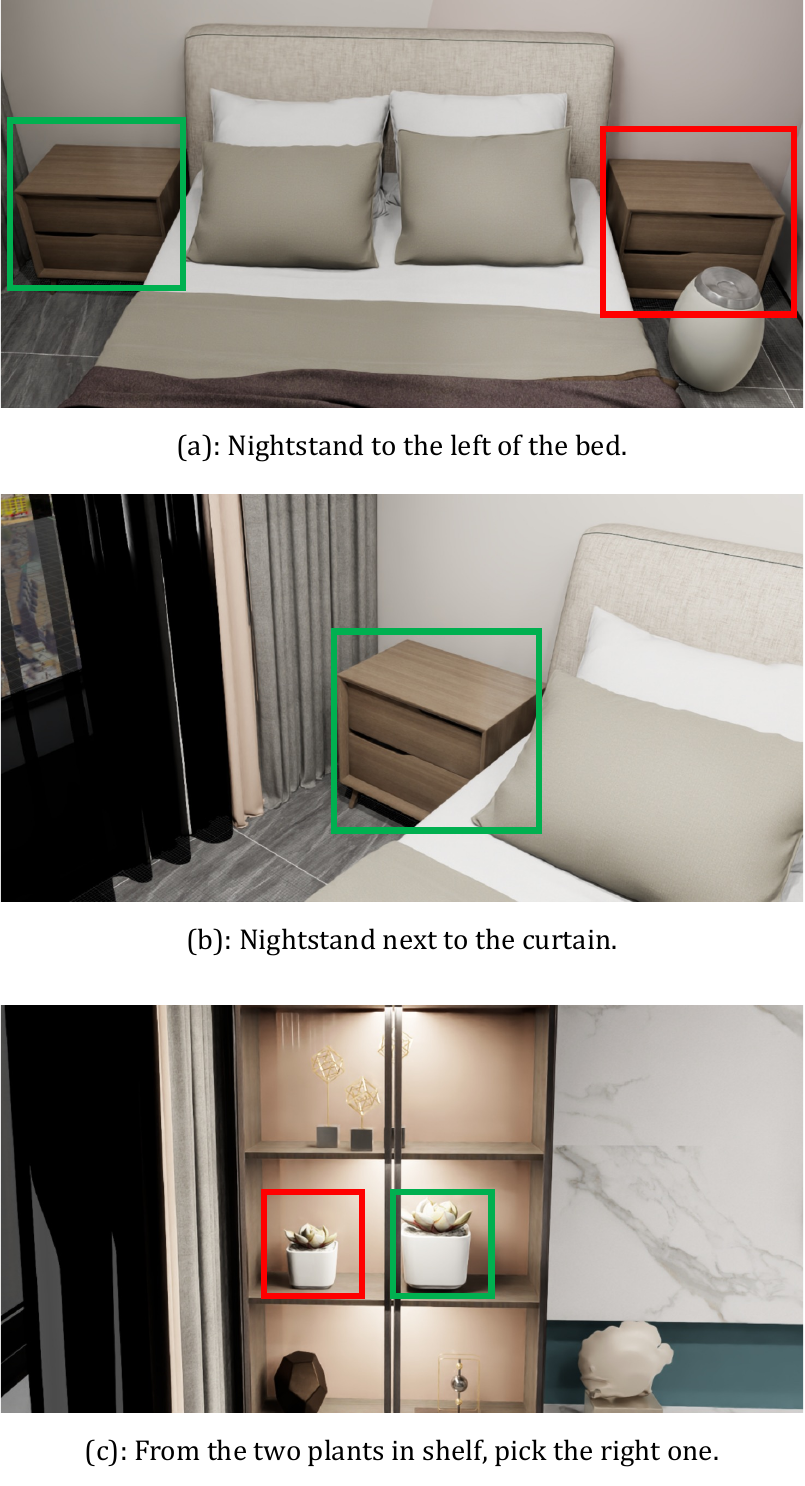} \hfill
  \includegraphics[width=0.48\linewidth]{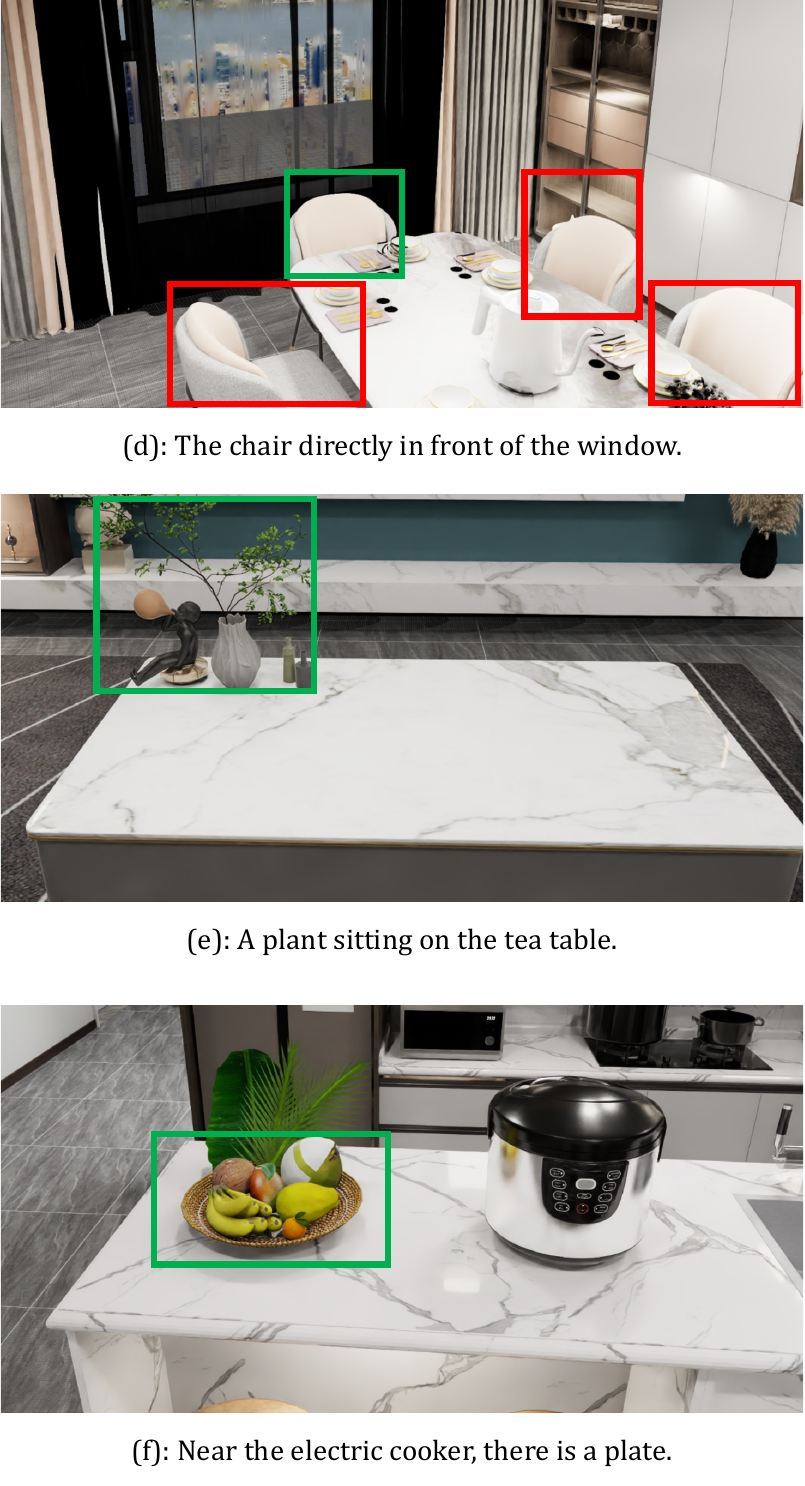}
  \caption {Qualitive results on GRScenes~\cite{grutopia}. The target object is in the green box and the visible distractors are indicated by the red box.}
  \label{fig:grscenes}
\end{figure*}

To validate the scene generalization of our relation, we select scenes from GRScenes~\cite{grutopia} and annotate relation-oriented referring utterances.
For evaluation, we directly use the object categories and bounding boxes.
Some examples of annotated data and results are shown in Figure~\ref{fig:grscenes}.

% \begin{figure*}
%     \centering
%     \includegraphics[width=\linewidth]{}
%     \caption{}
%     \label{fig:heatmap}
% \end{figure*}

\section{Public Resource Used}
\label{sec:ack}
In this section, we acknowledge the use of the following public resources for this work:

\begin{itemize}
    \item Pytorch \footnote{\url{https://github.com/pytorch/pytorch}} \dotfill Pytorch License 
    \item Referit3D \footnote{\url{https://github.com/referit3d/referit3d}} \dotfill MIT License
    \item GRScenes \footnote{\url{https://huggingface.co/datasets/OpenRobotLab/GRScenes}} \dotfill CC BY-SA 4.0 License
    \item ZSVG3D \footnote{\url{https://github.com/CurryYuan/ZSVG3D}} \dotfill Unknown
    \item Vil3drel \footnote{\url{https://github.com/cshizhe/vil3dref}} \dotfill Unknown
\end{itemize}

% \paragraph{Model Usage & Environmental Impact.}

% Codes are mainly generated with the GPT-4o~\cite{gpt4o} API. 
% The token 
% We have not trained or fine-tuned any additional large models.
\end{document}